\documentclass{article} 
\usepackage{iclr2027_conference,times}

\usepackage{amsmath,amsfonts,bm}

\def\eqref#1{equation~\ref{#1}}

\def\1{\bm{1}}

\DeclareMathAlphabet{\mathsfit}{\encodingdefault}{\sfdefault}{m}{sl}
\SetMathAlphabet{\mathsfit}{bold}{\encodingdefault}{\sfdefault}{bx}{n}

\usepackage{hyperref}
\usepackage{tabularx}
\usepackage{multirow}

\usepackage[per-mode=symbol, group-digits=integer, group-minimum-digits = 4, detect-all=true, input-symbols={()}, retain-explicit-plus]{siunitx}

\usepackage[percent]{overpic} 
\usepackage{fp} 
\usepackage{graphicx}
\usepackage{tikz}
\usetikzlibrary{positioning}
\usetikzlibrary{shapes.geometric, shapes, arrows, arrows.meta, bending}
\usetikzlibrary{backgrounds}
\usetikzlibrary{spy}
\usepackage{pgfplots}
\usepackage{bbding}
\usepackage{dsfont}
\usepackage{fontawesome5}
\usepackage{calc}
\pgfplotsset{compat=1.17}
\usepackage[outline]{contour} 
\contourlength{0.02em}
\usepackage[table]{xcolor}
\usepackage{makecell}
\usepackage{adjustbox}
\usepackage{pifont}
\newcommand{\cmark}{\ding{51}}%
\usepackage[scaled=0.97]{newtxtt}
\usepackage{utfsym}
\usepackage[super]{nth}
\newcommand\ours{FoRIS}

\usepackage{amsmath}

\newcommand{\eg}{e.g.}

\usepackage{cleveref}

\definecolor{cvprcolor}{RGB}{127,127,255}
\definecolor{softrow}{RGB}{245,245,245}
\definecolor{indomain}{gray}{0.55}
\newcommand{\ind}[1]{\textcolor{indomain}{#1}}

\definecolor{tud0d}{RGB}{83,83,83}
\definecolor{tud0c}{RGB}{137,137,137}
\definecolor{tud0b}{RGB}{181,181,181}
\definecolor{tud0a}{RGB}{220,220,220}
\definecolor{tud1a}{RGB}{93,133,195}
\definecolor{tud2a}{RGB}{0,156,218}
\definecolor{tud3a}{RGB}{80,182,149}
\definecolor{tud4a}{RGB}{175,204,80}
\definecolor{tud5a}{RGB}{221,223,72}
\definecolor{tud6a}{RGB}{255,224,92}
\definecolor{tud7a}{RGB}{248,186,60}
\definecolor{tud8a}{RGB}{238,122,52}
\definecolor{tud9a}{RGB}{233,80,62}
\definecolor{tud10a}{RGB}{201,48,142}
\definecolor{tud11a}{RGB}{128,69,151}
\definecolor{tud1b}{RGB}{0,90,169}
\definecolor{tud2b}{RGB}{0,131,204}
\definecolor{tud3b}{RGB}{0,157,129}
\definecolor{tud4b}{RGB}{153,192,0}
\definecolor{tud5b}{RGB}{201,212,0}
\definecolor{tud6b}{RGB}{253,202,0}
\definecolor{tud7b}{RGB}{245,163,0}
\definecolor{tud8b}{RGB}{236,101,0}
\definecolor{tud9b}{RGB}{230,0,26}
\definecolor{tud10b}{RGB}{166,0,132}
\definecolor{tud11b}{RGB}{114,16,133}
\definecolor{tud1c}{RGB}{0,78,138}
\definecolor{tud2c}{RGB}{0,104,157}
\definecolor{tud3c}{RGB}{0,136,119}
\definecolor{tud4c}{RGB}{127,171,22}
\definecolor{tud5c}{RGB}{177,189,0}
\definecolor{tud6c}{RGB}{215,172,0}
\definecolor{tud7c}{RGB}{210,135,0}
\definecolor{tud8c}{RGB}{204,76,3}
\definecolor{tud9c}{RGB}{185,15,34}
\definecolor{tud10c}{RGB}{149,17,105}
\definecolor{tud11c}{RGB}{97,28,115}
\definecolor{tud1d}{RGB}{36,53,114}
\definecolor{tud2d}{RGB}{0,78,115}
\definecolor{tud3d}{RGB}{0,113,94}
\definecolor{tud4d}{RGB}{106,139,55}
\definecolor{tud5d}{RGB}{153,166,4}
\definecolor{tud6d}{RGB}{174,142,0}
\definecolor{tud7d}{RGB}{190,111,0}
\definecolor{tud8d}{RGB}{169,73,19}
\definecolor{tud9d}{RGB}{156,28,38}
\definecolor{tud10d}{RGB}{115,32,84}
\definecolor{tud11d}{RGB}{76,34,106}

\definecolor{semantic}{RGB}{37, 150, 190}
\definecolor{part}{RGB}{102, 51, 1}
\definecolor{personalized}{RGB}{0, 0, 101}

\definecolor{myorange}{RGB}{255, 127, 15}
\definecolor{myblue}{RGB}{52, 153, 255}
\definecolor{mypurple}{RGB}{178, 102, 255}

\newcolumntype{C}[1]{>{\centering\arraybackslash}p{#1}}

\usepackage{wrapfig}

\usepackage{caption}

\usepackage{bbm}

\usepackage{ algorithm, algpseudocode}

\usepackage{url}

\definecolor{cvprblue}{rgb}{0.21,0.49,0.74}
\usepackage{booktabs}
\usepackage{pifont}
\usepackage[table]{xcolor}
\usepackage{subcaption} 
\usepackage{tikz}
\usetikzlibrary{calc}

\usepackage{array}      
\usepackage{makecell}   
\usepackage{multirow}
\usepackage{graphicx} 
\usepackage{textcomp} 
\usepackage{stfloats}
\usepackage{colortbl}

\newcommand{\myparagraph}[1]{\smallskip\noindent\textbf{#1}\hspace{0.4em}}
\newcommand{\myparagraphnospace}[1]{\smallskip\noindent\textbf{#1}}

\usepackage{xspace}

\newcommand{\cf}{cf.\xspace}

\title{FoRIS: Progressive Foreground Refinement for Training-Free In-Context Segmentation}

\author{
Ming Hu$^{1,2}$, Jianfu Yin$^{1,2}$, Mingyu Dou$^{1,2}$, Miaomiao Zhang$^{1,2}$, Yao Wang$^{3}$ \\
Cong Hu$^{4}$, Bingliang Hu$^{1}$, Quan Wang$^{1}$ \\
$^{1}$Xi'an Institute of Optics and Precision Mechanics, Chinese Academy of Sciences \\
$^{2}$University of Chinese Academy of Sciences \quad
$^{3}$Xi'an Jiaotong University \\
$^{4}$Zhongnan Hospital of Wuhan University \\
\texttt{huming708@gmail.com}
}

\iclrfinalcopy 
\begin{document}

\maketitle

\begin{figure}[h]
        \centering
    \includegraphics[width=\linewidth]{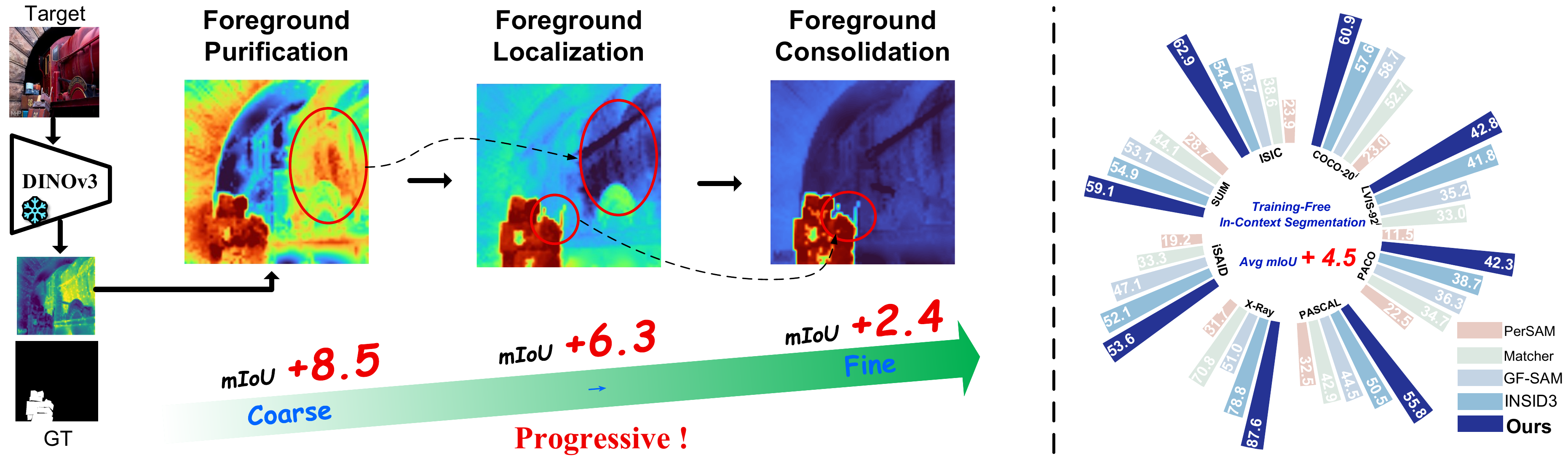}
    \caption{\textbf{Left:} Our core coarse-to-fine progressive refinement paradigm, which sequentially performs Foreground Purification, Foreground Localization, and Foreground Consolidation. Ablation results on COCO-20$^i$ show that each stage progressively improves the segmentation IoU. \textbf{Right:} Performance comparison with existing methods across diverse datasets spanning multiple domains, demonstrating the consistent effectiveness of FoRIS and its improvement in average mIoU.
 }
    \label{fig:method}
    \vspace{-0.2em}
\end{figure}

{\centering\large \textbf{Abstract}\par}

In-Context Segmentation (ICS) aims to precisely segment arbitrary semantic concepts, such as objects or parts, given one or a few annotated visual exemplars. In this paper, we revisit ICS from a more classical segmentation perspective, viewing it as a coarse-to-fine progressive refinement process. Rather than directly predicting the final mask through reference-query matching, we progressively refine the segmentation from coarse and ambiguous foreground responses to precise and complete foreground structures. Building upon this perspective, we propose a training-free in-context segmentation framework, termed FoRIS. Specifically, FoRIS consists of three key stages: Foreground Purification, Foreground Localization, and Foreground Consolidation, which progressively suppress background distractions, localize discriminative target regions, and recover complete foreground structures through semantic aggregation. Experimental results demonstrate that FoRIS achieves SOTA performance across semantic and part segmentation tasks, with average improvements of 4.5 and 4.8 mIoU points over existing approaches in the 1-shot and 5-shot settings, respectively.
Code: https://github.com/Xi-Mu-Yu/FoRIS.


\section{Introduction\label{sec:introduction}}

Segmentation is a fundamental task in computer vision with broad applications across natural, medical, and remote-sensing imagery~\cite{hu2025beta,hu2026fb,hu2024specslice,hu2026whereedit,ronneberger2015u,yang2023revisiting,yang2025unimatch,he2017mask,badrinarayanan2017segnet,chen2017deeplab,arbelaez2010contour,felzenszwalb2004efficient,achanta2012slic,shi2000normalized,long2015fully}. Traditional segmentation methods are typically restricted to predefined semantic categories and require substantial task-specific supervision, limiting their flexibility in open-world scenarios~\cite{cuttano2026insid3}. With the rapid development of visual foundation models (VFMs)~\cite{Caron:2021:DINO,Oquab:2023:Dinov2,Simeoni:2025:Dinov3,Kirillov:2023:SAM,carion2025sam}, \emph{in-context segmentation} (ICS) has emerged as a promising paradigm for open-world segmentation~\cite{cuttano2026insid3,Zhang:2023:PerSAM,Zhang:2024:GF-SAM,Liu:2023:Matcher}. Given one or more annotated reference examples at inference time, ICS aims to segment arbitrary target concepts without category-specific retraining, enabling flexible generalization across diverse concepts, domains, and annotation settings~\cite{cuttano2026insid3}.

Existing ICS methods generally follow two paradigms. \emph{Fine-tuned methods} adapt pretrained visual or diffusion architectures with task-specific segmentation supervision, translating pretrained representations into dense pixel-level predictions. Although effective within the training distribution, they require additional optimization and supervision, which can limit open-world generalization. In contrast, \emph{training-free methods} combine complementary pretrained models, typically leveraging visual representations for reference-query correspondence and segmentation priors for mask generation. While avoiding task-specific optimization, these methods still primarily rely on reference-query correspondence to transfer foreground semantics to the target image.

\begin{figure}
\centering
\includegraphics[width=\linewidth]{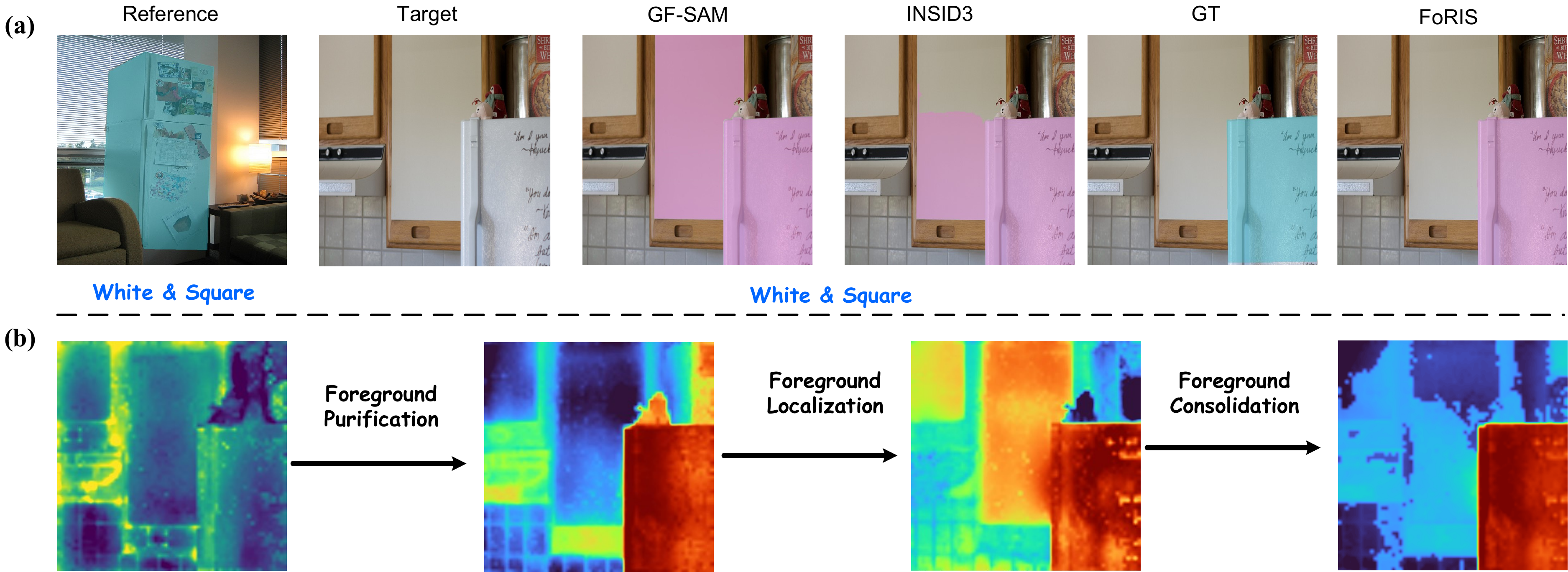}
\caption{\textbf{Motivation for progressive foreground refinement.}
(a) Existing training-free ICS methods rely on direct reference-query matching, which may confuse visually similar background regions with the target.
(b) We reinterpret ICS as a progressive foreground refinement process consisting of foreground purification, foreground localization, and foreground consolidation.}
\label{fig:motivation}
\vspace{-2em}
\end{figure}

Despite their methodological differences, existing training-free ICS methods
largely adopt a \emph{correspondence-centric} formulation, where foreground
semantics are transferred from reference exemplars to query images through
feature correspondence or similarity propagation~\cite{Liu:2023:Matcher,
Zhang:2024:GF-SAM,cuttano2026insid3}. As illustrated in Fig.~\ref{fig:motivation}(a),
given a reference image containing a white square-shaped refrigerator, such
correspondence-based methods may incorrectly associate visually similar
white square-shaped wall regions in the query image with the target, resulting
in inaccurate segmentation. This failure exposes a fundamental mismatch
between \emph{visual similarity} and \emph{target membership}: a query region
may closely resemble the reference foreground while being semantically
irrelevant to the target.

More fundamentally, an annotated foreground mask specifies \emph{where} the
target lies, but does not guarantee that the corresponding frozen features
provide a semantically pure representation of the target. Contextual,
positional, and background-correlated information may remain entangled with
foreground features. Meanwhile, variations in viewpoint, appearance, and
object structure can cause true target regions in the query image to produce
only partial or fragmented semantic responses. These factors introduce three
coupled sources of uncertainty: \emph{foreground contamination},
\emph{ambiguous target localization}, and \emph{fragmented foreground
responses}. Consequently, the central challenge of training-free ICS is not
simply to establish more accurate correspondence, but to progressively
transform noisy visual evidence into reliable, foreground-centric semantic
representations.

This observation motivates us to reconsider the role of correspondence in ICS. Rather than discarding correspondence, we view it as an intermediate source of semantic evidence within a \emph{progressive foreground refinement} process, as illustrated in Fig.~\ref{fig:motivation}(b). Specifically, foreground semantics progressively emerge through three stages: \emph{foreground purification}, \emph{foreground localization}, and \emph{foreground consolidation}. Foreground purification suppresses contextual and background interference in reference features, establishing a cleaner semantic basis for correspondence. Foreground localization then exploits the purified foreground cues to identify discriminative target regions in the query image, reducing localization ambiguity. Finally, foreground consolidation aggregates semantically related but spatially scattered responses, progressively recovering complete object structures from fragmented local evidence. In this view, correspondence is not treated as a sufficient endpoint, but as one component of a broader process that progressively refines semantic evidence into coherent object-level predictions.

Based on this perspective, we propose \textbf{FoRIS} (\textbf{Fo}reground \textbf{R}efinement for Training-Free \textbf{I}n-context \textbf{S}egmentation), a training-free ICS framework that explicitly implements this progressive foreground refinement paradigm. FoRIS sequentially performs foreground purification, foreground localization, and foreground consolidation using a frozen VFM, without additional segmentation supervision or auxiliary segmentation models. By placing correspondence within a progressive refinement pipeline, FoRIS shifts the focus from one-step semantic transfer toward the reliable extraction, localization, and consolidation of foreground evidence.

We further investigate a challenging and underexplored limitation of current VFM-based training-free ICS methods: their limited generalization to \emph{slender and fragmented structures}. Such structures often exhibit weak semantic responses, limited spatial support, and substantial appearance variations, amplifying the ambiguities that correspondence-based methods struggle to resolve. We systematically evaluate this challenge across diverse slender-structure datasets and find that existing training-free ICS methods remain substantially less reliable in such scenarios. These findings reveal a pronounced generalization gap despite the strong representations provided by VFMs, highlighting the need for training-free ICS methods that can accommodate heterogeneous target geometries and fragmented structures. Our contributions are summarized as follows:

\begin{itemize}
\item We introduce \emph{progressive foreground refinement} as a new perspective for training-free in-context segmentation, repositioning reference-query correspondence as an intermediate source of semantic evidence within a three-stage refinement process: foreground purification, localization, and consolidation.

\item We propose \textbf{FoRIS}, a training-free ICS framework that explicitly implements this progressive refinement paradigm and achieves average IoU improvements of over 4.5 and 4.8 percentage points over existing methods under the 1-shot and 5-shot settings, respectively.

\item We systematically investigate the generalization limitations of current VFM-based training-free ICS methods on slender and fragmented structures, revealing a challenging and underexplored scenario that remains an open problem for future ICS research.

\end{itemize}

\section{Related Work}
\label{sec:related}

\textbf{In-context segmentation.}
In-context segmentation (ICS) extends the in-context learning paradigm of LLMs~\cite{Brown:2020:GPT,Ouyang:2022:InstructGPT,Chowdhery:2023:Palm,Touvron:2023:Llama} to visual segmentation. Early works such as SegGPT~\cite{Wang:2023:SegGPT} and Painter~\cite{Wang:2023:Painter} demonstrated that segmentation can be conditioned on contextual examples, while recent approaches leverage visual foundation models (VFMs) for one-shot semantic, part, and personalized segmentation~\cite{Liu:2023:Matcher,Zhang:2023:PerSAM}. Unlike few-shot segmentation~\cite{Wang:2019:PaNet,Cuttano:2025:Sansa,Hong:2022:VAT,Lang:2022:BAM}, which typically targets predefined novel classes, ICS aims to segment arbitrary concepts specified at inference time across diverse semantic granularities.

Existing ICS methods can be broadly categorized into \emph{training-free} and \emph{supervised} approaches. Training-free methods~\cite{Liu:2023:Matcher,Zhang:2024:GF-SAM,Espinosa:2025:Notime,cuttano2026insid3} typically combine pretrained visual representations with external segmentation priors, while supervised methods~\cite{Meng:2024:SEGiC,Zhu:2024:Unleashing} adapt pretrained architectures through task-specific training. Despite their different implementations, existing approaches largely rely on reference-target correspondence as the primary mechanism for transferring foreground semantics and inferring target masks. However, correspondence alone may be insufficient when reference features are contaminated by background information or target responses are ambiguous and fragmented.
Rather than discarding correspondence, we reinterpret its role within a \emph{progressive foreground refinement} process. Inspired by the coarse-to-fine nature of segmentation, our approach progressively purifies foreground representations, localizes discriminative target regions, and consolidates fragmented semantic responses, allowing correspondence-derived evidence to evolve into coherent object masks.

\myparagraphnospace{Progressive Segmentation and Mask Refinement.}
A central challenge in training-free segmentation is to transform the rich
visual representations provided by vision foundation models (VFMs) into
reliable foreground-background separation. This objective is closely related
to a long-standing principle in classical segmentation: accurate object masks
often emerge through \emph{progressive refinement} rather than a single
prediction step. Classical approaches progressively reduce segmentation
uncertainty by updating region assignments, object boundaries, or spatial
consistency. Region Growing~\cite{adams1994seeded} expands initial seeds
according to local similarity, while Graph Cut~\cite{boykov2001interactive}
and GrabCut~\cite{rother2004grabcut} iteratively optimize
foreground-background assignments through global region relationships.
Active Contour~\cite{kass1988snakes} progressively evolves object boundaries,
and multi-scale approaches~\cite{adelson1984pyramid} refine object structures
from coarse to fine resolutions. Despite their methodological differences,
these approaches share a common principle: reliable segmentation emerges by
progressively transforming uncertain local evidence into coherent foreground
structures.

In contrast, existing training-free ICS methods largely adopt a
\emph{correspondence-centric} formulation, where foreground semantics are
transferred from annotated reference images to target images through feature
correspondence or similarity propagation~\cite{Liu:2023:Matcher,
Zhang:2024:GF-SAM,cuttano2026insid3}. Although some methods further employ
clustering, propagation, or mask refinement, these operations are primarily
built upon correspondence-derived evidence rather than explicitly modeling
foreground refinement as the central inference process. Consequently, they
remain vulnerable to background interference, ambiguous semantic
correspondence, and fragmented target responses, which may lead to incomplete
or disconnected predictions and semantic drift. This observation motivates
us to reinterpret training-free ICS as a \emph{progressive foreground
refinement} process, where correspondence serves as intermediate semantic
evidence that is progressively purified, localized, and consolidated into
coherent object-level predictions.

\section{In-context Segmentation with FoRIS}
\label{sec:method}

Given reference image--mask pairs $\{(I_s, M_s)\}_{s=1}^{S}$ and a target image $I_t$, FoRIS predicts $\hat{M}_t$ without fine-tuning or auxiliary heads.
We extract patch features from the last intermediate layer of the encoder and $\ell_2$-normalize them along the channel dimension:
\begin{equation}
    \mathbf{F} = \phi\big([I_1, \ldots, I_S, I_t]\big) \in \mathbb{R}^{(S{+}1) \times C \times H \times W},
    \qquad \mathbf{f}(\mathbf{p}) = \mathbf{F}(\mathbf{p}) / \|\mathbf{F}(\mathbf{p})\|_2.
\end{equation}
FoRIS refines a foreground response through three stages---\emph{foreground purification} (FP), \emph{foreground localization} (FL), and \emph{foreground consolidation} (FC).
\cref{alg:foris} summarizes the full data flow.

\subsection{Foreground Purification}
\label{sec:foreground_purification}

FP removes positional layout bias and background clutter via adaptive positional debiasing (APD) and two-stage foreground refinement (FR).

\paragraph{Adaptive Positional Debiasing.}
Positional bias can hinder prototype matching when reference and target objects occupy inconsistent locations.
However, in domains with strong spatial regularity---\eg, lung CT, where anatomical structures follow consistent layouts---positional cues can serve as informative priors rather than noise.
We therefore debias \emph{adaptively} instead of indiscriminately.
Following INSID3~\cite{cuttano2026insid3}, we offline estimate a low-rank positional subspace $\mathbf{U}$ via SVD on centered features from a zero-input image.
At inference, we measure reference--target semantic alignment:
\begin{equation}
    s_{\mathrm{sem}} =
    \frac{\langle \boldsymbol{\mu}^{\mathrm{fg}}, \bar{\mathbf{f}}_t \rangle}{\|\bar{\mathbf{f}}_t\|_2},
    \qquad
    \boldsymbol{\mu}^{\mathrm{fg}} = \mathrm{norm}\!\Big(
        \mathrm{mean}_{\mathbf{p} \in \Omega^{\mathrm{fg}}} \mathbf{f}_r(\mathbf{p})
    \Big),
\end{equation}
where $\bar{\mathbf{f}}_t = \mathrm{mean}_{\mathbf{p}} \mathbf{f}_t(\mathbf{p})$.
Debiasing is applied to \emph{all} reference and target tokens only when $s_{\mathrm{sem}} < \theta$:
\begin{equation}
    \mathbf{P}_\perp = \mathbf{I}_C - \mathbf{U}\mathbf{U}^\top, \qquad
    \tilde{\mathbf{f}}(\mathbf{p}) =
    \begin{cases}
        \mathrm{norm}\big(\mathbf{P}_\perp \mathbf{f}(\mathbf{p})\big), & s_{\mathrm{sem}} < \theta, \\[2pt]
        \mathbf{f}(\mathbf{p}), & \text{otherwise}.
    \end{cases}
\label{eq:apd}
\end{equation}
We denote the resulting feature maps as $\tilde{\mathbf{F}}$ and use $\tilde{\mathbf{f}}$ for all subsequent stages unless stated otherwise.
We set $\theta{=}0.80$ by default (\cf \cref{sec:ablation_fp,fig:debias_threshold_sweep}).

\begin{figure}
        \centering
    \includegraphics[width=\linewidth]{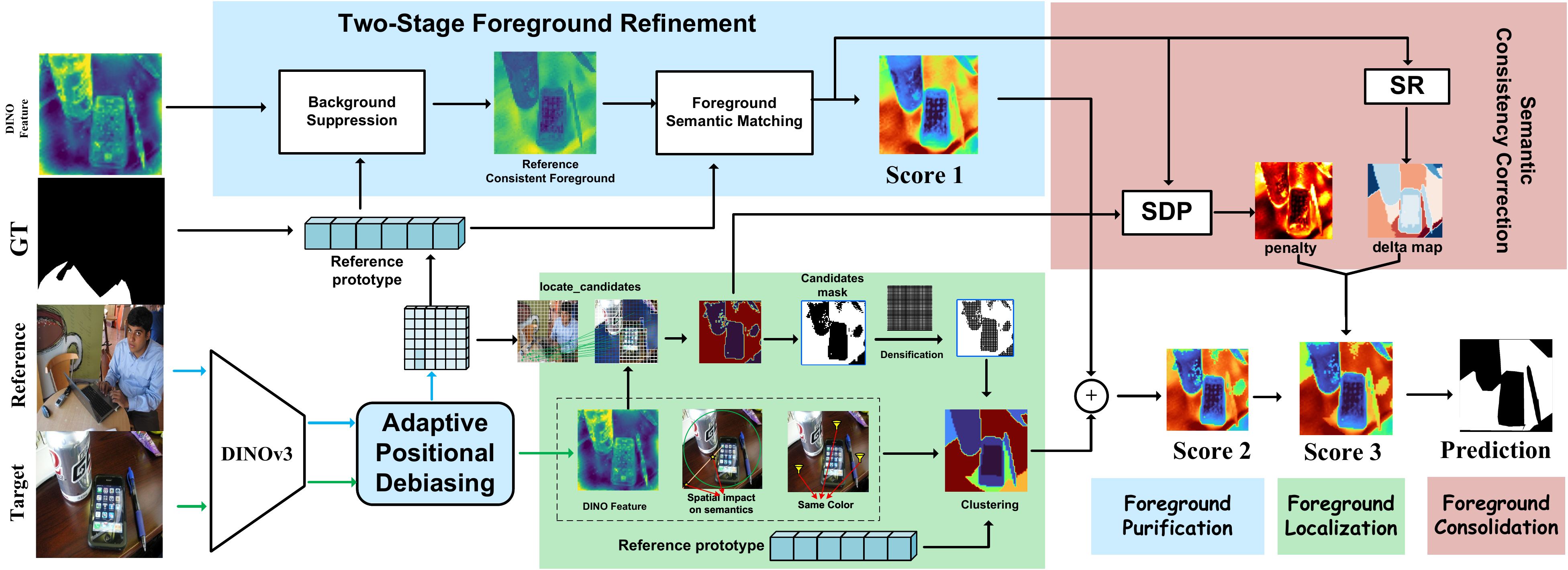}
    \caption{\textbf{Overview of \ours{}.} Given reference image(s) with foreground mask(s) and a target image, \ours{} performs through three sequential stages. \textbf{(1) Foreground purification (FP):} patch features from a frozen DINOv3 encoder are adaptively debiased and purified via two-stage foreground refinement, including contrastive background suppression and foreground semantic matching, yielding an initial response (\emph{Score~1}). \textbf{(2) Foreground localization (FL):} bidirectional feature matching identifies reliable foreground candidates, which are densified and integrated with clustering-based seed priors built from DINO, spatial, and color cues to produce a refined response (\emph{Score~2}). \textbf{(3) Foreground consolidation (FC):} semantic disagreement penalization (SDP) and semantic reweighting (SR) suppress conflicting activations and strengthen semantically pure regions, producing the final response (\emph{Score~3}).}
    \label{fig:method}
    \vspace{-0.2em}
\end{figure}

\paragraph{Two-stage foreground refinement.}
FR operates on debiased features $\tilde{\mathbf{f}}$ and produces both an initial response and a gated feature map used later in FC.

\emph{Stage~1 (feature gating).}
We first construct foreground and background prototypes from debiased reference tokens.
Foreground tokens from all references are concatenated and partitioned by average-linkage agglomerative clustering under cosine distance, yielding $J$ cluster prototypes $\boldsymbol{\mu}_j = \mathrm{norm}\!\big(\mathrm{mean}_{i : \ell_i = j} \mathbf{x}_i^{\mathrm{fg}}\big)$ for $j = 1, \ldots, J$, where $J$ is determined by the clustering procedure rather than fixed manually and $\ell_i$ is the cluster label of foreground token $i$.
The hard-negative background prototype targets background patches most confusable with the foreground.
We rank all debiased background tokens by their cosine similarity to $\boldsymbol{\mu}^{\mathrm{fg}}$ and retain the top 20\%:
\begin{equation}
    \mathcal{I}_{\mathrm{hn}} = \mathrm{TopK}_{20\%}\!\Big(
        \big\{ \langle \mathbf{x}_i^{\mathrm{bg}}, \boldsymbol{\mu}^{\mathrm{fg}} \rangle \big\}_{i=1}^{N_{\mathrm{bg}}}
    \Big),
\end{equation}
and define $\boldsymbol{\mu}^{\mathrm{bg}} = \mathrm{norm}\!\big(\mathrm{mean}_{i \in \mathcal{I}_{\mathrm{hn}}} \mathbf{x}_i^{\mathrm{bg}}\big)$.
A contrastive gate is then computed on all reference and target tokens and applied multiplicatively.
We aggregate similarities to the $J$ foreground cluster prototypes via temperature-scaled log-sum-exp:
\begin{equation}
    g(\mathbf{p}) = \sigma\!\Big(
        \beta \log \sum_{j=1}^{J} \exp\!\Big(
            \frac{\langle \tilde{\mathbf{f}}(\mathbf{p}), \boldsymbol{\mu}_j \rangle}{\beta}
        \Big)
        - \langle \tilde{\mathbf{f}}(\mathbf{p}), \boldsymbol{\mu}^{\mathrm{bg}} \rangle
    \Big), \qquad
    \hat{\mathbf{f}}(\mathbf{p}) = g(\mathbf{p}) \cdot \tilde{\mathbf{f}}(\mathbf{p}),
\label{eq:gate}
\end{equation}
where $\beta{=}0.07$ and reference foreground locations receive a gate floor.
Stage~1 yields gated features $\hat{\mathbf{f}}$, which are passed to FC for cluster reweighting (\cref{sec:foreground_consolidation}).

\emph{Stage~2 (foreground semantic matching).}
Score~1 is computed on \emph{debiased but ungated} target features $\tilde{\mathbf{f}}_t$, using prototypes rebuilt from debiased reference tokens.
This decouples contrastive gating from prototype matching: gating suppresses background-dominated activations for consolidation, while matching preserves the full debiased semantic geometry for cross-image correspondence.
The orthogonalized background direction is
\begin{equation}
    \boldsymbol{\mu}^{\mathrm{bg}}_\perp =
    \frac{
        \boldsymbol{\mu}^{\mathrm{bg}} -
        \langle \boldsymbol{\mu}^{\mathrm{bg}}, \boldsymbol{\mu}^{\mathrm{fg}} \rangle \boldsymbol{\mu}^{\mathrm{fg}}
    }{
        \|\boldsymbol{\mu}^{\mathrm{bg}} -
        \langle \boldsymbol{\mu}^{\mathrm{bg}}, \boldsymbol{\mu}^{\mathrm{fg}} \rangle \boldsymbol{\mu}^{\mathrm{fg}}\|_2
    },
\label{eq:orth_bg}
\end{equation}
and the initial purified response is
\begin{equation}
    S^{(1)}(\mathbf{p}) =
    \beta \log \sum_{j=1}^{J} \exp\!\Big(
        \frac{\langle \tilde{\mathbf{f}}_t(\mathbf{p}), \boldsymbol{\mu}_j \rangle}{\beta}
    \Big)
    - \langle \tilde{\mathbf{f}}_t(\mathbf{p}), \boldsymbol{\mu}^{\mathrm{bg}}_\perp \rangle.
\label{eq:score1}
\end{equation}

\subsection{Foreground Localization}
\label{sec:foreground_localization}

FL refines $S^{(1)}$ into Score~2 using debiased reference and target features $\tilde{\mathbf{f}}$.

\paragraph{Cross-image candidate voting.}
Dense patch-to-patch matching assigns a foreground vote to each target location:
\begin{equation}
    V(\mathbf{p}) = \frac{1}{S}\sum_{s=1}^{S}
    \mathbbm{1}\!\Big[
        M_s\!\big(\arg\max_{\mathbf{q}} \langle \tilde{\mathbf{f}}_s(\mathbf{q}), \tilde{\mathbf{f}}_t(\mathbf{p}) \rangle\big)
    \Big].
\end{equation}
Locations with majority support form $\mathcal{C} = \{\mathbf{p} \mid V(\mathbf{p}) > \tfrac{1}{2}\}$, which is lightly densified on a coarse spatial grid.

\paragraph{Multi-cue clustering and seed-cluster prior.}
Target patches are clustered on a joint representation of $\ell_2$-normalized DINO features, RGB color, and 2D coordinates:
\begin{equation}
    \mathbf{z}_i = \mathrm{norm}\big([\mathbf{f}_i;\, \mathbf{c}_i;\, \mathbf{p}_i]\big).
\label{eq:multicue}
\end{equation}
The same agglomerative clustering protocol as above assigns each patch a cluster id $\kappa(\mathbf{p}) \in \{1,\ldots,J'\}$.
Let $\mathcal{G}_j = \{\mathbf{p} : \kappa(\mathbf{p}) = j\}$ denote the patch set of cluster $j$.
If no cluster overlaps $\mathcal{C}$, we set $P(\mathbf{p}) = \mathbbm{1}[\mathbf{p} \in \mathcal{C}]$; otherwise, the seed cluster $j^\star$ maximizes reference affinity weighted by candidate area:
\begin{equation}
    j^\star = \arg\max_{j \in \mathcal{J}_{\mathcal{C}}}
    \underbrace{\mathrm{mean}_{\mathbf{p} \in \mathcal{G}_j}\langle \tilde{\mathbf{f}}_t(\mathbf{p}), \boldsymbol{\mu}^{\mathrm{fg}} \rangle}_{\text{cross-image affinity}}
    \cdot
    \underbrace{\frac{|\mathcal{G}_j \cap \mathcal{C}|}{|\mathcal{G}_j|}}_{\text{area weight}},
\end{equation}
where $\mathcal{J}_{\mathcal{C}} = \{j : \mathcal{G}_j \cap \mathcal{C} \neq \emptyset\}$.
All clusters receive
\begin{equation}
    \rho_j =
    \mathrm{mean}_{\mathbf{p} \in \mathcal{G}_j}\langle \tilde{\mathbf{f}}_t(\mathbf{p}), \boldsymbol{\mu}^{\mathrm{fg}} \rangle
    \cdot \max\big(\langle \boldsymbol{\mu}_{j^\star}, \boldsymbol{\mu}_j \rangle, 0\big)
    \cdot \frac{|\mathcal{G}_j \cap \mathcal{C}|}{|\mathcal{G}_j|},
\end{equation}
which is min--max normalized across clusters and mapped to patches as
\begin{equation}
    P(\mathbf{p}) = \frac{\rho_{\kappa(\mathbf{p})} - \min_i \rho_i}{\max_i \rho_i - \min_i \rho_i + \epsilon}
    \in [0,1].
\end{equation}
Score~2 is
\begin{equation}
    S^{(2)} = S^{(1)} + (V - \tfrac{1}{2}) + (P - \tfrac{1}{2}).
\label{eq:score2}
\end{equation}

\subsection{Foreground Consolidation}
\label{sec:foreground_consolidation}

FC transforms Score~2 into Score~3 via semantic disagreement penalization (SDP) and semantic reweighting (SR).

\paragraph{Semantic disagreement penalization.}
We penalize inter-map disagreement and foreground--background coupling:
\begin{equation}
    \mathcal{D}(\mathbf{p}) = |S_{\mathrm{fg}}(\mathbf{p}) - V(\mathbf{p})| + |S_{\mathrm{fg}}(\mathbf{p}) - P(\mathbf{p})|,
    \qquad
    \mathcal{B}(\mathbf{p}) = \min\big(S_{\mathrm{fg}}(\mathbf{p}), S_{\mathrm{bg}}(\mathbf{p})\big).
\end{equation}
An uncertainty-modulated penalty concentrates corrections on ambiguous patches:
\begin{equation}
    \Pi(\mathbf{p}) = \big(\mathcal{D}(\mathbf{p}) + \mathcal{B}(\mathbf{p})\big)
    \cdot \mathcal{U}(\mathbf{p}), \qquad
    \mathcal{U}(\mathbf{p}) = 1 - 2\big|S_{\mathrm{fg}}(\mathbf{p}) - \tfrac{1}{2}\big|.
\label{eq:sdp}
\end{equation}

\paragraph{Semantic reweighting.}
Target patches are re-clustered on gated features $\hat{\mathbf{f}}_t$, yielding a new cluster assignment $\kappa'(\mathbf{p})$.
Each cluster $j$ receives a reweighting factor based on foreground--background separation:
\begin{equation}
    \Delta_j =
    \big(\overline{S}_{\mathrm{fg}}^{(j)} - \overline{S}_{\mathrm{bg}}^{(j)}\big)_+
    - \min\big(\overline{S}_{\mathrm{fg}}^{(j)}, \overline{S}_{\mathrm{bg}}^{(j)}\big),
\end{equation}
where $\overline{S}_{\cdot}^{(j)}$ is the mean over $\{\mathbf{p} : \kappa'(\mathbf{p}) = j\}$.
The patch-level map is $\Delta(\mathbf{p}) = \Delta_{\kappa'(\mathbf{p})}$, and the consolidated response is
\begin{equation}
    S^{(3)} = S^{(2)} - \Pi + \Delta.
\label{eq:score3}
\end{equation}

\section{Experiments}
\label{sec:experiments}

We evaluate \ours{} on one-shot \emph{semantic} and \emph{part} segmentation.
Following INSID3~\cite{cuttano2026insid3}, we adopt the same datasets, baselines, and evaluation protocols for fair comparison.
All results are reported as mIoU; implementation details are provided in Appendix~\cref{sec:experiments_appendix}.



\subsection{Main Results}

\begin{table}[t]
\caption{
\textbf{Comparison of \ours{} (mIoU in \%, $\uparrow$) on one-shot semantic, part, and personalized segmentation.} 
State-of-the-art methods are grouped into {task-specific fine-tuning} and {training-free} approaches. 
Previous training-free methods rely on SAM, pre-trained with mask-level supervision, whereas \ours{} uses only frozen self-supervised DINOv3 features. 
\textcolor{indomain}{Gray} indicates the model was trained on the corresponding train split of the dataset; best results \textbf{bold}, \nth{2} best \underline{underlined}.
}
\label{tab:main_comparison}

\vspace{-2mm}

\centering
\small
\setlength{\tabcolsep}{3pt}

\resizebox{\linewidth}{!}{
\begin{tabular}{@{}lcc|cccccc|cc|cc@{}}
\toprule

& & &
\multicolumn{6}{c}{\textbf{Semantic}} &
\multicolumn{2}{c}{\textbf{Part}} & \\

\textbf{Method} &
\textbf{Encoder} &
\textbf{\#Param} &
\textbf{LVIS-92} &
\textbf{COCO-20} &
\textbf{ISIC} &
\textbf{SUIM} &
\textbf{iSAID} &
\textbf{X-Ray} &
\textbf{PASCAL} &
\textbf{PACO} &
\textbf{Avg}
\\

\midrule

\multicolumn{12}{@{}l}{\textbf{Task-specific fine-tuning}: \textit{Semantic + mask supervision}} \\
~~Painter \cite{Wang:2023:Painter}
& \small{ViT} & \SI{354}{M}
& 10.5 & \ind{33.1} & -- & -- & -- & -- & 30.4 & 14.1 & -- \\

~~SegGPT \cite{Wang:2023:SegGPT}
& \small{ViT} & \SI{354}{M}
& 18.6 & \ind{56.1} & 37.5 & 34.9 & \ind{30.9}
& \underline{87.5} & 35.8 & \ind{13.5}
& 39.4 \\

~~SINE \cite{Liu:2024:SINE}
& \small{DINOv2} & \SI{373}{M}
& 31.2 & \ind{64.5} & 25.8 & 50.7 & 38.3 & 39.8 & 36.2 & 23.3
& 38.7 \\

~~DiffewS \cite{Zhu:2024:Unleashing}
& \small{Stable Diffusion} & \SI{890}{M}
& 31.4 & \ind{71.3} & 27.8 & 48.9 & 47.5 & 41.6 & 34.0 & 22.8
& 40.7 \\

~~SegIC \cite{Meng:2024:SEGiC}
& \small{DINOv2} & \SI{310}{M}
& \ind{44.6} & \ind{76.1}
& 25.3 & 52.5 & 46.1 & 34.5 & 39.9 & 25.9
& 43.1 \\

~~SegIC\,{\scriptsize (COCO)}\,\cite{Meng:2024:SEGiC}
& \small{DINOv2} & \SI{310}{M}
& 35.7 & \ind{75.6}
& 22.5 & 52.9 & 40.8 & 30.8 & 38.6 & 25.1
& 40.3 \\

\midrule

\multicolumn{12}{@{}l}{\textbf{Training free}: \textit{Mask-supervised pre-training}} \\
~~PerSAM \cite{Zhang:2023:PerSAM}
& \small{SAM} & \SI{640}{M}
& 11.5 & 23.0 & 23.9 & 28.7 & 19.2 & 31.7 & 32.5 & 22.5
& 24.1 \\

~~Matcher \cite{Liu:2023:Matcher}
& \small{DINOv2 + SAM} & \SI{945}{M}
& 33.0 & 52.7 & 38.6 & 44.1 & 33.3 & 70.8 & 42.9 & 34.7
& 43.7 \\

~~GF-SAM \cite{Zhang:2024:GF-SAM}
& \small{DINOv2 + SAM} & \SI{945}{M}
& 35.2 & \underline{58.7} & 48.7 & 53.1 & 47.1 & 51.0 & 44.5 & 36.3
& 46.8 \\

GF-SAM$^\dagger$ \cite{Zhang:2024:GF-SAM}
& DINOv3+SAM & 945M
& 31.8 & 54.8 & 50.9 & 50.5 & 46.7
& 56.1 & 44.9 & 34.4
& 46.3
\\

\midrule

\multicolumn{12}{l}{
\textbf{Training-free}: \textit{Unsupervised pre-training}
}
\\

INSID3
& DINOv3 & 304M
& \underline{41.8}
& 57.6
& \underline{54.4}
& \underline{54.9}
& \underline{52.1}
& 78.8
& \underline{50.5}
& \underline{38.7}
& \underline{53.6}
\\

\textbf{FoRIS (ours)}
& DINOv3 & \textbf{304M}
& \textbf{42.8}
& \textbf{60.9}
& \textbf{62.9}
& \textbf{59.1}
& \textbf{53.6}
& \textbf{87.6}
& \textbf{55.8}
& \textbf{42.3}
& \textbf{58.1}
\\

\bottomrule
\end{tabular}
}

\vspace{-2mm}
\end{table}
\begin{figure*}
        \centering
    \includegraphics[width=\linewidth]{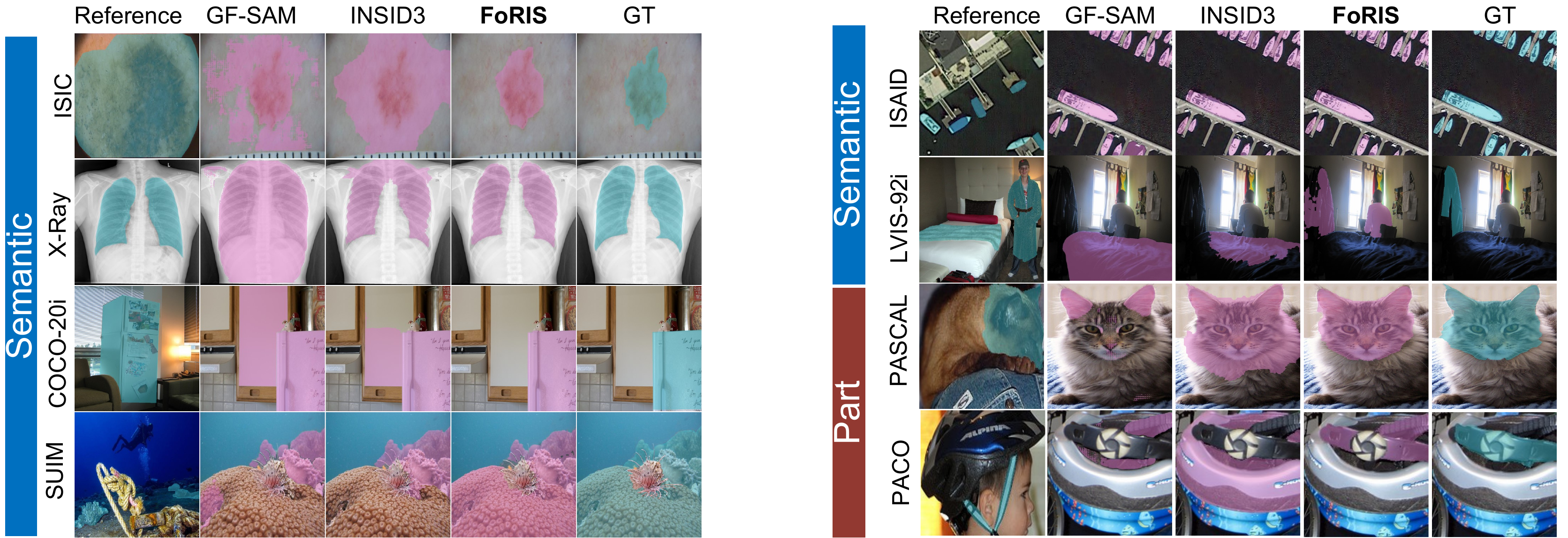}
    \vspace{-1.75em}
\caption{Comparison of \textbf{\ours{}} with GF-SAM~\cite{Zhang:2024:GF-SAM} and INSID3~\cite{cuttano2026insid3} on one-shot semantic and part segmentation.}
    \label{fig:qualitatives}
    \vspace{-0.5em}
\end{figure*}

\textbf{One-shot semantic segmentation.}
\ours{} achieves the best performance on all semantic benchmarks (\cf \cref{tab:main_comparison}).
Compared with the strongest training-free baseline GF-SAM, \ours{} improves mIoU by \SI{7.6}{\%}, \SI{2.2}{\%}, \SI{14.2}{\%}, \SI{6.0}{\%}, \SI{6.5}{\%}, and \SI{36.6}{\%} pts.\ on LVIS-92$^i$, COCO-20$^i$, ISIC, SUIM, iSAID, and Chest X-ray, with the largest gains on out-of-domain datasets.
Under the same frozen DINOv3 backbone and parameter budget, \ours{} consistently surpasses INSID3 by \SI{1.0}{\%}--\SI{8.8}{\%} pts.\ across all six benchmarks.
\ours{} also generalizes better than task-specific fine-tuned methods without any segmentation training.
Qualitative results in \cref{fig:qualitatives}  show that \ours{} produces clean, semantically coherent masks directly from frozen features.

\textbf{One-shot part segmentation.}
\ours{} also sets a new SOTA on part-level benchmarks.
It outperforms GF-SAM by \SI{11.3}{\%} and \SI{6.0}{\%} pts.\ on PASCAL-Part and PACO-Part, and INSID3 by \SI{5.3}{\%} and \SI{3.6}{\%} pts., respectively.
Compared with fine-tuned SegIC and DiffewS, \ours{} achieves up to \SI{21.8}{\%} pts.\ higher mIoU while remaining fully training-free.
These results indicate that progressive foreground refinement benefits fine-grained structural understanding beyond object-level segmentation.
Qualitative comparisons in \cref{fig:qualitatives} confirm improved boundary quality and part integrity.

\subsection{Ablation Study}
\label{sec:ablation}

We conduct ablation experiments to validate each component of \ours{} (\cf \cref{tab:ablation}).
Starting from the baseline, progressively adding FP, FL, and FC consistently improves performance across all benchmarks, confirming the complementarity of the three-stage design.

\begin{wraptable}[18]{r}{0.5\textwidth}
\centering
\caption{Ablation study of different modules.}
\label{tab:ablation}

\vspace{0mm}
\setlength{\tabcolsep}{1.5pt}
\renewcommand{\arraystretch}{1.15}

\small
\begin{tabular}{ccccc|cccc}
\toprule
\multirow{2}{*}{\textbf{Base}} &
\multicolumn{2}{c}{\textbf{FP}} &
\multirow{2}{*}{\textbf{FL}} &
\multirow{2}{*}{\textbf{FC}} &
\multicolumn{4}{c}{\textbf{mIoU}}\\

\cmidrule(lr){2-3}
\cmidrule(lr){6-9}

&
\textbf{APD} &
\textbf{FR} &
&
&
\textbf{COCO} &
\textbf{ISIC} &
\textbf{SUIM} &
\textbf{PASCAL}
\\

\midrule

\rowcolor{gray!10}
\cmark & & & & &
43.7 & 46.8 & 44.4 & 38.7 \\

\cmark & \cmark & & & &
46.9 & 41.2 & 43.9 & 39.3 \\

\cmark & & \cmark & & &
48.3 & 57.1 & 50.2 & 44.4 \\

\cmark & & & \cmark & &
51.9 & 53.1 & 53.2 & 51.8 \\

\midrule

\rowcolor{gray!20}
\cmark & \cmark & \cmark & & &
52.2 & 58.7 & 52.9 & 48.3 \\

\cmark & \cmark & & \cmark & &
52.9 & 49.3 & 55.6 & 50.8 \\

\rowcolor{gray!20}
\cmark & & \cmark & \cmark & &
56.2 & 58.0 & 55.8 & 53.8 \\

\cmark & \cmark & \cmark & \cmark & &
58.5 & 60.2 & 58.9 & 55.1 \\

\rowcolor{gray!20}
\cmark & & \cmark & \cmark & \cmark &
58.0 & 59.2 & 54.8 & 54.2 \\

\rowcolor{blue!25}
\cmark & \cmark & \cmark & \cmark & \cmark &
\textbf{60.9} & \textbf{62.9} &
\textbf{59.1} & \textbf{55.8} \\

\bottomrule
\end{tabular}

\vspace{-2mm}
\end{wraptable}

\subsubsection{Foreground Purification}
\label{sec:ablation_fp}

\begin{figure}[t]
    \centering
    \includegraphics[width=\linewidth]{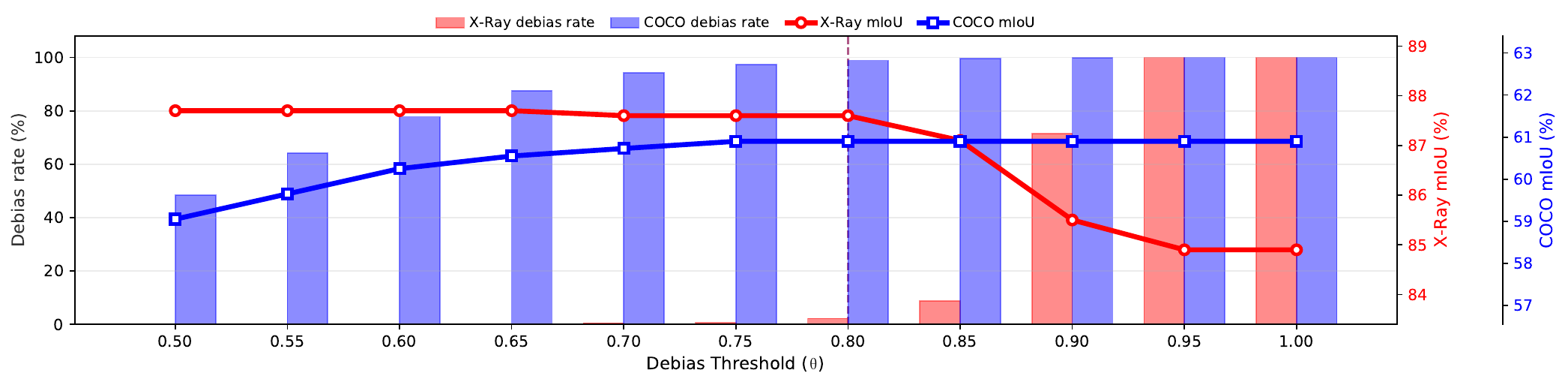}
    \vspace{-2em}
\caption{Effect of the adaptive debiasing threshold $\theta$ on COCO and Chest X-ray.
}
    \label{fig:debias_threshold_sweep}
\end{figure}

We examine the dual role of positional bias: it can interfere with correspondence when reference and target layouts differ, yet provide useful structural priors in domains with stable spatial organization, such as Chest X-ray. APD therefore adaptively applies debiasing only when the semantic alignment score $s_{\mathrm{sem}} < \theta$, projecting features onto the orthogonal complement of the positional subspace.
We sweep $\theta \in \{0.50, 0.55, \ldots, 1.00\}$ on COCO and Chest X-ray (\cref{fig:debias_threshold_sweep}). On COCO, mIoU increases from \SI{59.1}{\%} to \SI{60.9}{\%} and saturates at $\theta \ge 0.75$, indicating that suppressing positional bias benefits cross-layout matching. In contrast, Chest X-ray remains stable at \SI{87.6}{\%}--\SI{87.7}{\%} for $\theta \le 0.80$ but drops to \SI{84.9}{\%} at $\theta{=}1.0$, showing that excessive debiasing can remove useful anatomical priors.

\cref{tab:ablation} further confirms the dataset-dependent role of APD: enabling APD alone improves COCO by \SI{3.2}{\%} pts.\ ($43.7 \rightarrow 46.9$) but reduces ISIC by \SI{5.6}{\%} pts.\ ($46.8 \rightarrow 41.2$). With all modules enabled, FoRIS achieves \SI{60.9}{\%}/\SI{62.9}{\%}/\SI{59.1}{\%}/\SI{55.8}{\%} mIoU on COCO/ISIC/SUIM/PASCAL-Part, with up to \SI{2.9}{\%} pts. improvement over the best variant without APD. These results support \emph{adaptive}, rather than always-on, positional debiasing for cross-domain generalization.

\subsubsection{Foreground Localization}
\label{sec:ablation_fl}

\textbf{Multi-cue clustering.}
We study whether augmenting frozen DINOv3 features with positional coordinates and RGB appearance during clustering improves FL (\cf \cref{tab:rgb_pos_results}).
Positional cues benefit Chest X-ray (\SI{+1.1}{\%} pts.), where anatomical structures follow strong spatial regularity, while RGB cues help SUIM (\SI{+0.2}{\%} pts.), where appearance variation is more pronounced.
Combining both cues maintains strong performance across domains (\SI{87.6}{\%} on Chest X-ray, \SI{59.1}{\%} on SUIM), and we adopt this joint representation by default.

\textbf{Dense candidate aggregation.}
We further evaluate coarse-grid densification of the candidate set (\cf \cref{tab:dense}).
Dense aggregation consistently improves mIoU on COCO (\SI{60.1}{\%} $\rightarrow$ \SI{60.9}{\%}), Chest X-ray (\SI{87.4}{\%} $\rightarrow$ \SI{87.6}{\%}), and PACO-Part (\SI{41.8}{\%} $\rightarrow$ \SI{42.3}{\%}).
The gains are largest on fine-grained PACO-Part, where richer local context helps disambiguate subtle part-level differences.

\begin{table*}[t]
\centering

\caption{Ablation studies of different components in \ours{}.}
\label{tab:all_ablation}

\vspace{0em}

\begin{subfigure}{0.48\linewidth}
\centering
\caption{Comparison between original and adaptive positional debiasing on X-Ray, COCO, and PACO.}
\label{tab:debias_ablation}

\renewcommand{\arraystretch}{1.15}

\begin{tabular}{c|ccc}
\toprule
\textbf{Method}
& \textbf{X-Ray} $\uparrow$
& \textbf{COCO} $\uparrow$
& \textbf{PACO} $\uparrow$ \\
\midrule
OD & 84.9 & 60.9 & 42.3 \\
AD & \textbf{87.6} & 60.9 & 42.3 \\
\bottomrule
\end{tabular}
\end{subfigure}
\hfill
\begin{subfigure}{0.48\linewidth}
\centering
\caption{Ablation study of densification in the FL module on COCO, X-Ray, and PACO.}
\label{tab:dense}

\setlength{\tabcolsep}{8pt}
\renewcommand{\arraystretch}{1.15}

\begin{tabular}{c c c c}
\toprule
\textbf{Method} & \textbf{COCO} $\uparrow$ & \textbf{X-Ray} $\uparrow$ & \textbf{PACO} $\uparrow$ \\
\midrule
w/o Dense & 60.1 & 87.4 & 41.8 \\
w/ Dense & \textbf{60.9} & \textbf{87.6} & \textbf{42.3} \\
\bottomrule
\end{tabular}
\end{subfigure}

\vspace{0.5em}

\begin{subfigure}{0.48\linewidth}
\centering
\caption{Ablation study of positional and RGB cues in the FL module on X-Ray and SUIM.}
\label{tab:rgb_pos_results}

\setlength{\tabcolsep}{8pt}
\renewcommand{\arraystretch}{1.15}

\begin{tabular}{c c c}
\toprule
\textbf{Method} & \textbf{X-Ray} $\uparrow$ & \textbf{SUIM} $\uparrow$ \\
\midrule
None & 86.7 & 59.0 \\
Pos & \textbf{87.8} & 58.8 \\
RGB & 86.8 & \textbf{59.2} \\
Pos + RGB & 87.6 & 59.1 \\
\bottomrule
\end{tabular}
\end{subfigure}
\hfill
\begin{subfigure}{0.48\linewidth}
\centering
\caption{Ablation study of consolidation components in the FC module on ISIC, COCO, and PACO.}
\label{tab:FC_ab}

\setlength{\tabcolsep}{6pt}
\renewcommand{\arraystretch}{1.15}

\begin{tabular}{c c c c}
\toprule
\textbf{Method} & \textbf{ISIC} $\uparrow$ & \textbf{COCO} $\uparrow$ & \textbf{PACO} $\uparrow$ \\
\midrule
None & 60.2 & 58.5 & 39.8 \\
SDP & 61.1 & 59.9 & 41.7 \\
SR & 62.6 & 60.2 & 40.9 \\
SDP + SR & \textbf{62.9} & \textbf{60.9} & \textbf{42.3} \\
\bottomrule
\end{tabular}
\end{subfigure}

\end{table*}

\subsubsection{Foreground Consolidation}
\label{sec:ablation_fc}

\cref{tab:FC_ab} evaluates the two FC components: semantic disagreement penalization (SDP) and semantic reweighting (SR).
SDP consistently improves the baseline by suppressing unreliable activations caused by inter-map disagreement and foreground--background coupling, with the largest gain on PACO-Part (\SI{+1.9}{\%} pts.).
SR provides stronger improvements on ISIC and COCO by correcting region-level semantic bias through cluster-wise reweighting.
Combining SDP and SR achieves the best performance on all benchmarks, confirming their complementarity.

\section{Challenges and the future of VFM-based Training-Free ICS}


\begin{wrapfigure}[15]{r}{0.4\linewidth}
    \centering

    \includegraphics[width=\linewidth]{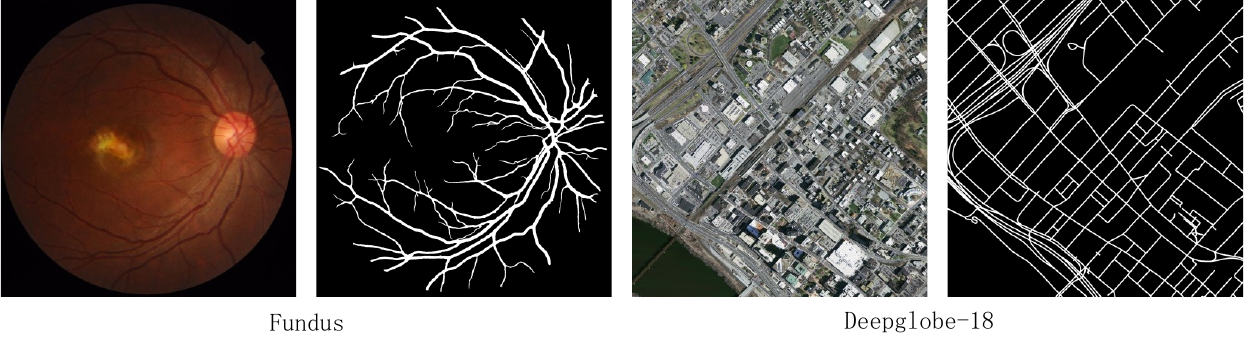}

    \caption{Visualization of the Fundus and DeepGlobe-18 datasets.}
    \label{fig:slender_challenge}


    \captionof{table}{Comparison of results between SAM-based and DINO-based methods.}
    \label{tab:slender_challenge}

    \setlength{\tabcolsep}{3pt}
    \renewcommand{\arraystretch}{1.15}

    \small
    \begin{tabular}{l|cc|cc}
    \toprule

    \multirow{2}{*}{\textbf{Method}}
    &
    \multicolumn{2}{c|}{\textbf{Fundus}}
    &
    \multicolumn{2}{c}{\textbf{DeepGlobe 18}}
    \\

    \cmidrule(lr){2-3}
    \cmidrule(lr){4-5}

    &
    \textbf{1-shot}
    &
    \textbf{5-shot}
    &
    \textbf{1-shot}
    &
    \textbf{5-shot}
    \\

    \midrule

    GF-SAM
    & 8.6 & 8.4
    & 8.5 & 9.3
    \\

    INSID3
    & 12.3 & 11.7
    & 7.2 & 7.3
    \\

    \textbf{\ours{}}
    & \textbf{19.7} & \textbf{21.5}
    & \textbf{13.3} & \textbf{13.5}
    \\

    \bottomrule
    \end{tabular}

\end{wrapfigure}

        \vspace{-2mm}




















When extending ICS to the slender-structure scenarios illustrated in Fig.~\ref{fig:slender_challenge}, Tab.~\ref{tab:slender_challenge} shows that existing SAM- and DINO-based ICS methods achieve consistently low IoU on Fundus and DeepGlobe-18, revealing their limitations in slender-structure segmentation. This raises two questions: why are slender structures particularly challenging for VFM-based training-free ICS, and how can segmentation performance be improved in such scenarios? Further analysis and experiments are provided in Appendix~\ref{challenge_apendix}.

\section{Conclusion}


In this work, inspired by the coarse-to-fine paradigm in classical image processing, we rethink the fundamental formulation of ICS and propose a foreground progressive ICS framework, \textbf{FoRIS}. Rather than treating ICS as a direct matching problem between reference and target images, we reformulate it as a \emph{progressive foreground refinement} process, which progressively suppresses background interference, localizes target regions, and recovers coherent object structures through three stages: foreground purification, foreground localization, and foreground consolidation.
Extensive experiments demonstrate that FoRIS achieves competitive performance across semantic segmentation, part segmentation, and cross-domain benchmarks, while exhibiting strong out-of-distribution generalization. Meanwhile, we reveal the limitations of existing training-free ICS methods on \emph{slender and fragmented structures}: weak semantic responses, limited spatial support, and substantial appearance variations make direct feature matching inadequate for preserving structural continuity. How to fully exploit VFM representations to handle such weakly salient, fine-grained, and topology-sensitive targets remains an important open problem for training-free ICS.

\clearpage

\bibliography{bibtex/short, bibtex/references}
\bibliographystyle{iclr2027_conference}

\clearpage

\appendix

\section*{Appendix}

\section{Experiments setting}
\label{sec:experiments_appendix}
We evaluate \ours{} on one-shot \emph{semantic} and \emph{part} segmentation.
In each setting, a single annotated reference mask is provided, and the model is tasked with segmenting the corresponding concept in the target image:
\emph{(1) semantic} -- all instances of a given class (\eg, “dog”);
\emph{(2) part} -- same object part (\eg, “dog ear”).

For \textbf{one-shot \emph{semantic} segmentation}, we use six datasets across a range of imaging scenarios: COCO-20$^i$~\cite{Nguyen:2019:Feature} with 80 object categories; LVIS-92$^i$~\cite{Liu:2023:Matcher} with 920 categories and a strong long-tail distribution; 
ISIC2018~\cite{Codella:2019:Skin, Tschandl:2018:Ham10000} for skin lesion segmentation; Chest X-Ray~\cite{Candemir:2013:Lung, Jaeger:2013:Automatic}, an X-ray dataset of lung screening; iSAID-5$^i$~\cite{Yao:2021:Isaid}, a remote sensing dataset with 15 categories; and SUIM~\cite{Islam:2020:Suim} with underwater imagery and 8 categories. For \textbf{one-shot \emph{part} segmentation}, we use PASCAL-Part~\cite{Liu:2023:Matcher}, providing 56 object parts across 15 categories, and PACO-Part~\cite{Liu:2023:Matcher} with 303 object parts from 75 categories. 

\myparagraph{Implementation details.}
We adopt the Large version of the DINOv3~\cite{Simeoni:2025:Dinov3} encoder. 
Input images are resized to 1024 $\times$ 1024, following SAM-based approaches~\cite{Liu:2023:Matcher, Zhang:2023:PerSAM, Zhang:2024:GF-SAM}. The final segmentation masks are predicted at patch resolution: we bilinearly interpolate them to original resolution, and apply mask refinement with a CRF \cite{Krahebuhl:2011:CRF}, following \cite{Hamilton:2022:USS, Hahn:2025:CUPS, Gansbeke:2021:Unsupervised, Melas:2022:Spectral}.

\section{Algorithm Summary}

\begin{algorithm}[htbp]
\caption{FoRIS inference (single target episode)}
\label{alg:foris}
\begin{algorithmic}[1]
\Require Reference images $\{I_s, M_s\}_{s=1}^{S}$, target image $I_t$, frozen encoder $\phi$
\Ensure Binary mask $\hat{M}_t$
\State $\mathbf{F} \gets \phi([I_1,\ldots,I_S,I_t])$; $\mathbf{f}(\mathbf{p}) \gets \mathbf{F}(\mathbf{p})/\|\mathbf{F}(\mathbf{p})\|_2$
\State \textbf{// FP: adaptive positional debiasing}
\State Compute $s_{\mathrm{sem}}$; if $s_{\mathrm{sem}} < \theta$, replace $\mathbf{f}$ with $\tilde{\mathbf{f}}$ via $\mathbf{P}_\perp$ (\cref{eq:apd})
\State \textbf{// FP: Stage-1 gating}
\State Cluster reference foreground tokens (agglomerative) $\to$ $\{\boldsymbol{\mu}_j\}_{j=1}^{J}$; build $\boldsymbol{\mu}^{\mathrm{fg}}$, $\boldsymbol{\mu}^{\mathrm{bg}}$ (TopK 20\% hard negatives)
\State Compute gate $g(\mathbf{p})$ and gated features $\hat{\mathbf{f}}(\mathbf{p})$ on all tokens (\cref{eq:gate})
\State \textbf{// FP: Stage-2 matching (Score~1)}
\State Orthogonalize $\boldsymbol{\mu}^{\mathrm{bg}}_\perp$ w.r.t.\ $\boldsymbol{\mu}^{\mathrm{fg}}$ (\cref{eq:orth_bg})
\State $S^{(1)} \gets \beta \log \sum_j \exp(\langle \tilde{\mathbf{f}}_t, \boldsymbol{\mu}_j \rangle / \beta) - \langle \tilde{\mathbf{f}}_t, \boldsymbol{\mu}^{\mathrm{bg}}_\perp \rangle$ on \emph{ungated} $\tilde{\mathbf{f}}_t$ (\cref{eq:score1})
\State $S_{\mathrm{fg}}, S_{\mathrm{bg}} \gets$ per-episode min--max normalization
\State \textbf{// FL (Score~2)}
\State $V \gets$ cross-image candidate votes; densify $\mathcal{C}$
\State Cluster target patches on $[\mathbf{f}; \mathbf{c}; \mathbf{p}]$ (\cref{eq:multicue}); build seed prior $P$ from $j^\star$ and $\{\rho_j\}$
\State $S^{(2)} \gets S^{(1)} + (V-\tfrac{1}{2}) + (P-\tfrac{1}{2})$ (\cref{eq:score2})
\State \textbf{// FC (Score~3)}
\State $\Pi \gets$ SDP from $S_{\mathrm{fg}}, S_{\mathrm{bg}}, V, P$ (\cref{eq:sdp})
\State Re-cluster gated $\hat{\mathbf{f}}_t$; compute $\Delta(\mathbf{p})$ from cluster means
\State $S^{(3)} \gets S^{(2)} - \Pi + \Delta$ (\cref{eq:score3})
\State $\hat{M}_t \gets$ binarize, upsample, CRF-refine $S^{(3)}$
\State \Return $\hat{M}_t$
\end{algorithmic}
\end{algorithm}

\section{Semantic Correspondence Analysis of Positional Debiasing}
\label{app:debias_correspondence}

\begin{table}[htbp]
\centering
\small
\caption{\textbf{Semantic correspondence on SPair-71k}
(PCK@$T$ in \%, $\uparrow$). Comparison across DINOv3
backbones with original, debiased, and adaptive debias variants.}
\label{tab:debias_spair}
\vspace{-0.5em}

\resizebox{\linewidth}{!}{%
\begin{tabular}{
@{}c
c
ccc
ccc
ccc
@{}
}
\toprule

&& \multicolumn{3}{c}{\textbf{Small}}
& \multicolumn{3}{c}{\textbf{Base}}
& \multicolumn{3}{c}{\textbf{Large}} \\

\cmidrule(lr{0.8em}){3-5}
\cmidrule(lr{0.8em}){6-8}
\cmidrule(lr){9-11}

{$T$} &&
{\textit{original}} &
{\textit{debias}} &
{\textbf{\textit{adaptive}}} &

{\textit{original}} &
{\textit{debias}} &
{\textbf{\textit{adaptive}}} &

{\textit{original}} &
{\textit{debias}} &
{\textbf{\textit{adaptive}}} \\

\midrule

\bfseries 0.05 &&
27.93 & 28.18 & \bfseries 28.37 &
30.07 & 33.68 & \bfseries 33.69 &
33.55 & 34.57 & \bfseries 34.61 \\

\bfseries 0.10 &&
44.92 & 45.65 & \bfseries 45.94 &
46.80 & 52.60 & \bfseries 52.60 &
52.04 & 54.11 & \bfseries 54.12 \\
 
\bfseries 0.15 &&
54.31 & 55.75 & \bfseries 56.01 &
55.60 & 62.49 & \bfseries 62.50 &
61.63 & 64.24 & \bfseries 64.27 \\

\bfseries 0.20 &&
60.55 & 62.78 & \bfseries 62.97 &
61.22 & 68.72 & \bfseries 68.73 &
67.80 & 70.76 & \bfseries 70.76 \\

\bottomrule
\end{tabular}
}

\vspace{-0.65em}
\end{table}

To further investigate the effect of positional debiasing on semantic correspondence, we evaluate the original, fixed-debiasing, and adaptive-debiasing variants on SPair-71k. The results are reported in \cref{tab:debias_spair}.

Applying positional debiasing consistently improves semantic correspondence across different DINOv3 model scales and evaluation thresholds. For example, with the DINOv3-Base backbone, PCK@0.10 improves from 46.80 to 52.60, while PCK@0.20 increases from 61.22 to 68.73. Similar improvements are observed across the Small, Base, and Large backbones, indicating that removing position-dominated components can improve the reliability of dense semantic correspondence.

Adaptive debiasing further achieves the strongest performance across the evaluated settings. Although its improvement over fixed debiasing is relatively modest, the adaptive variant consistently avoids the performance degradation observed when positional information is suppressed indiscriminately. This suggests that positional information in frozen DINOv3 features is not purely harmful: useful spatial regularities can facilitate correspondence when reference and target images are already well aligned.

This observation is also consistent with the segmentation ablation in \cref{tab:ablation}. Applying debiasing unconditionally does not always improve downstream segmentation and can even degrade performance on some datasets, such as ISIC (46.8$\rightarrow$41.2). In contrast, adaptive debiasing dynamically regulates the debiasing strength according to the reliability of semantic correspondence, suppressing position-dominated activations only when necessary.

Overall, the SPair-71k results provide complementary evidence for the design of APD: effective positional debiasing should selectively suppress harmful positional bias while retaining useful geometric priors encoded in frozen visual features.

\section{Computational Cost of FoRIS}

\label{sec:computational_analysis}

We further analyze the computational cost of \ours{} on COCO-20$^i$ from both the component-wise and resolution-wise perspectives. As shown in Table~\ref{tab:computationaldetailed}, the total inference time of \ours{} is \SI{1620.77}{ms} for a single in-context example at a resolution of \SI{1024}{px} on a single RTX 3090. The main computational costs come from the encoder forward pass and the subsequent foreground localization and consolidation stages, while foreground purification introduces only a relatively small overhead. Although \ours{} is slower than INSID3~\cite{cuttano2026insid3} at the same \SI{1024}{px} resolution, directly comparing runtime at a fixed resolution does not fully characterize the accuracy--efficiency trade-off between the two methods.

To further investigate this trade-off, we evaluate \ours{} under different input resolutions on COCO-20$^i$, as reported in Table~\ref{tab:computationalresolution}. Interestingly, \ours{} achieves \SI{59.5}{\%} mIoU even at a resolution of only \SI{512}{px}, already surpassing the \SI{57.6}{\%} mIoU achieved by INSID3 at \SI{1024}{px}. More importantly, the \SI{512}{px} setting requires only \SI{258}{ms} per in-context example, compared with \SI{812}{ms} for INSID3 at \SI{1024}{px}. This corresponds to a \SI{68.2}{\%} reduction in inference time while simultaneously improving mIoU by \SI{1.9}{percentage points}. Thus, although \ours{} incurs a higher computational cost when operating at the same resolution, it can achieve stronger segmentation performance with substantially lower-resolution inputs, yielding a more favorable practical accuracy--efficiency trade-off on COCO-20$^i$.

The resolution study further reveals a clear diminishing-return effect. Increasing the input resolution of \ours{} from \SI{512}{px} to \SI{1024}{px} improves mIoU from \SI{59.5}{\%} to \SI{60.9}{\%}, yielding only a \SI{1.4}{percentage-point} gain, whereas the inference time increases from \SI{258}{ms} to \SI{1621}{ms}, corresponding to more than a $6\times$ increase in computational cost. Further increasing the resolution provides little additional benefit: mIoU remains around \SI{61}{\%} even at \SI{1760}{px}, while the inference time increases substantially to \SI{11235}{ms}. These results indicate that, on COCO-20$^i$, the segmentation performance of \ours{} is relatively insensitive to increasing input resolution, and that low-resolution inference can already preserve most of its segmentation capability.

Overall, the results on COCO-20$^i$ demonstrate that the computational efficiency of \ours{} should be considered jointly with its achievable segmentation accuracy at different resolutions, rather than solely through runtime comparisons at a fixed resolution. In particular, the \SI{512}{px} setting provides a compelling operating point, achieving higher mIoU than INSID3 at \SI{1024}{px} while requiring only a fraction of its inference time. This favorable accuracy--efficiency trade-off suggests that \ours{} can maintain strong in-context segmentation performance without relying on computationally expensive high-resolution inputs.

\begin{table}[htbp]
\centering
\small
\renewcommand{\arraystretch}{1.0}
\setlength{\tabcolsep}{2pt}
\caption{\textbf{Detailed computational analysis of \ours{} on COCO-20$^i$.}
We report the inference time of each component in milliseconds (ms, $\downarrow$) for a single in-context example.
Runtime is measured at a resolution of \SI{1024}{px} on a single RTX 3090.}
\vspace{-0.4em}
\begin{tabularx}{\linewidth}{@{}Xr}
\toprule
\textbf{Component} & \textbf{Runtime} $\downarrow$ \\
\midrule
Encoder forward & \SI{702.27}{ms} \\
Foreground purification & \SI{38.74}{ms} \\
Foreground localization & \SI{346.44}{ms} \\
Foreground consolidation & \SI{342.50}{ms} \\
CRF refinement & \SI{190.82}{ms} \\
\midrule
\textbf{Total inference time} & \bfseries \SI{1620.77}{ms} \\
\midrule
\textbf{INSID3~\cite{cuttano2026insid3}} & \bfseries \SI{812}{ms} \\
\bottomrule
\end{tabularx}
\vspace{-0.3em}
\label{tab:computationaldetailed}
\end{table}

\begin{table}[htbp]
    \centering
    \setlength{\tabcolsep}{1pt}
    \small
   \caption{\textbf{Resolution-wise computational analysis of \ours{} on COCO-20$^i$.}
    We report the inference time and mIoU of \ours{} under different input resolutions.
    Runtime is measured for a single in-context example on a single RTX 3090.
    The \SI{1024}{px} setting corresponds to the resolution used for the component-wise analysis in Table~\ref{tab:computationaldetailed}.}
    \vspace{-0.4em}

    \begin{tabularx}{\linewidth}{@{}X X X X X X X X@{}}
        \toprule
        \multicolumn{7}{c}{\textit{Low resolution} $\leftarrow$ \textbf{Resolution} $\rightarrow$ \textit{High resolution}} \\
        \midrule

        \bfseries
        &
        \bfseries\SI{512}{px} 
        & \bfseries\SI{720}{px} 
        & \bfseries\SI{896}{px} 
        & \bfseries\SI{1024}{px} 
        & \bfseries\SI{1440}{px} 
        & \bfseries\SI{1600}{px} 
        & \bfseries\SI{1760}{px}
        \\

        \midrule

        \textbf{Runtime}
        &
        \SI{258}{ms}
        &
        \SI{593}{ms}
        &
        \SI{1108}{ms}
        &
        \SI{1621}{ms}
        &
        \SI{5468}{ms}
        &
        \SI{9018}{ms}
        &
        \SI{11235}{ms}
        \\

        \textbf{mIoU (\%)}
        &
        59.5
        &
        60.3
        &
        60.7
        &
        60.9
        &
        60.8
        &
        60.1
        &
        61.0
        \\

        \bottomrule
    \end{tabularx}%

    \label{tab:computationalresolution}
    \vspace{-0.3em}
\end{table}

\section{n-shot segmentation}

We further evaluate \ours{} in the 5-shot setting by providing five reference image--mask pairs for each target image. As shown in Table~\ref{tab:kshot}, \ours{} achieves the highest average mIoU of \textbf{64.7\%}, outperforming the strongest training-free baseline, INSID3, by \textbf{4.8 percentage points}. \ours{} ranks first on \textbf{7 out of 8 benchmarks} and consistently outperforms INSID3 across all datasets, with particularly notable gains on X-Ray (\textbf{+7.9 points}), PASCAL (\textbf{+6.2 points}), and PACO (\textbf{+5.4 points}). These improvements span diverse domains, including natural images, medical images, underwater scenes, and aerial imagery, demonstrating the robustness of \ours{} under substantial domain shifts. Notably, \ours{} uses the same DINOv3 encoder and parameter count as INSID3, indicating that the performance gains arise from more effective foreground purification, localization, and consolidation rather than increased model capacity. Moreover, all hyperparameters are directly reused from the 1-shot setting without any additional tuning, demonstrating that \ours{} scales effectively to multiple contextual examples.

\begin{table*}[htbp]
\caption{
\textbf{Comparison of \ours{} (mIoU in \%, $\uparrow$) on 5-shot semantic and part segmentation.} Models are provided with 5 contextual examples and tasked with segmenting the annotated concept in the target image. 
\ours{} scales effectively to multiple references, achieving robust performance across domains. 
All hyperparameters are reused from the 1-shot setting without any tuning, highlighting the versatility of our approach.
\textcolor{indomain}{Gray} indicates the model was trained on the corresponding train split of the dataset; best results \textbf{bold}, \nth{2} best \underline{underlined}.
}
\label{tab:kshot}
\vspace{-2mm}
\centering
\small
\setlength{\tabcolsep}{2.5pt}

\resizebox{\linewidth}{!}{
\begin{tabular}{lcc|cccccc|cc|c}
\toprule
\textbf{Method} &
\textbf{Encoder} &
\textbf{Params} &
\multicolumn{6}{c|}{\textbf{Semantic}} &
\multicolumn{2}{c|}{\textbf{Part}} &
\textbf{Avg}
\\
&
&
&
LVIS-92$_i$ &
COCO-20$_i$ &
ISIC &
SUIM &
iSAID &
X-Ray &
PASCAL &
PACO &
\\
\midrule

\multicolumn{12}{l}{\textit{Task-specific fine-tuning: semantic + mask supervision}}
\\

SegGPT &
ViT &
354M &
25.4 &
\ind{67.9} &
45.2 &
33.7 &
\ind{35.9} &
\textbf{89.1} &
42.8 &
\ind{14.1} &
44.3
\\

SINE &
DINOv2 &
373M &
35.5 &
\ind{66.1} &
28.6 &
54.8 &
40.5 &
40.6 &
36.4 &
25.4 &
41.0
\\

DiffewS &
Stable Diffusion &
890M &
35.4 &
\ind{72.2} &
32.7 &
49.8 &
48.0 &
45.1 &
39.7 &
26.1 &
43.6
\\

\midrule

\multicolumn{12}{l}{\textit{Training-free: mask-supervised pre-training}}
\\

Matcher &
DINOv2+SAM &
945M &
40.0 &
60.7 &
35.0 &
50.6 &
34.3 &
71.2 &
45.8 &
33.6 &
46.4
\\

GF-SAM &
DINOv2+SAM &
945M &
\underline{44.2} &
\underline{66.8} &
55.2 &
58.1 &
52.4 &
52.9 &
51.2 &
\underline{41.9} &
52.8
\\

GF-SAM$^\dagger$ &
DINOv3+SAM &
945M &
42.8 &
64.4 &
56.7 &
58.6 &
53.2 &
54.7 &
50.3 &
39.3 &
52.5
\\

GF-SAM$^\dagger$ + debias &
DINOv3+SAM &
945M &
43.6 &
64.6 &
{58.2} &
{59.2} &
{54.1} &
59.1 &
{51.4} &
40.2 &
{53.8}
\\

\midrule

\multicolumn{12}{l}{\textit{Training-free: unsupervised pre-training}}
\\

INSID3 &
DINOv3 &
\textbf{304M} &
\underline{47.2} &
65.1 &
\underline{63.9} &
\underline{61.7} &
\underline{56.9} &
80.1 &
\underline{57.1} &
\underline{46.8} &
\underline{59.9}
\\

\textbf{\ours} &
DINOv3 &
\textbf{304M} &
\textbf{51.0} &
\textbf{68.4} &
\textbf{66.5} &
\textbf{66.9} &
\textbf{61.7} &
\underline{88.0} &
\textbf{63.3} &
\textbf{52.2 } &
\textbf{64.7}
\\

\bottomrule
\end{tabular}
}

\vspace{-3mm}
\end{table*}

\section{Limitations of VFM-based Training-Free ICS for Slender Structure Segmentation}
\label{challenge_apendix}

Vision foundation models (VFMs) have emerged as powerful visual representation learners and have substantially advanced training-free in-context segmentation (ICS).
By leveraging large-scale pre-trained knowledge, existing VFM-based ICS methods perform segmentation without task-specific optimization, typically by measuring feature similarity between support and query images, constructing semantic prototypes, or exploiting correspondence encoded in foundation-model representations.
These approaches achieve strong results on objects with distinct semantics and relatively compact spatial extent.

However, the suitability of such training-free paradigms for slender-structure segmentation remains uncertain.
Unlike common objects with large, coherent regions, slender structures---blood vessels, nerves, and tubular anatomical patterns---exhibit properties that conflict with the assumptions underlying current VFM-based ICS.
Segmenting them requires not only semantic recognition but also precise localization, boundary preservation, and structural continuity, none of which are explicitly optimized in standard VFM representations.

\paragraph{Patch-level foreground--background mixing.}
Most foundation models, especially transformer-based architectures, represent images with spatial tokens derived from local patches.
For compact objects, individual tokens usually contain enough foreground signal to capture object semantics.
For slender structures, however, foreground may occupy only a small fraction of a patch, so the resulting representation is dominated by surrounding background.
The extracted feature therefore mixes foreground and background rather than encoding a pure object descriptor.
This weakens separability between slender structures and their context and makes reliable training-free matching difficult.

\paragraph{Dispersed intra-class feature distributions.}
Training-free ICS methods often assume that instances of the same category form a consistent, compact cluster in feature space.
Prototype-based approaches, for instance, summarize foreground regions into representative embeddings and label query regions by similarity to these prototypes.
Slender structures violate this assumption: scale, orientation, curvature, illumination, and local appearance vary widely along a single structure, and different segments can look substantially different.
The foreground distribution in feature space becomes dispersed rather than compact, so a single prototype may fail to cover the full appearance and topology of an elongated target.

\subsection{Experimental Setup}
\label{sec:slender_setup}

To empirically examine the limitations above, we treat slender-structure segmentation as a \emph{stress test} for training-free ICS rather than as a solved benchmark.
We evaluate on two public datasets with elongated, low-contrast foreground:
Fundus~\cite{jin2022fives} for retinal vessel segmentation and DeepGlobe-18~\cite{demir2018deepglobe} for aerial road extraction.
Both exhibit low appearance saliency, elongated topology, and weak local discriminability.
We follow the episodic protocol of the main experiments: each episode contains $S$ reference image--mask pairs and one target image, and all methods use a frozen DINOv3-L encoder without task-specific fine-tuning.
Unless otherwise noted, inputs are resized to $1024{\times}1024$ and features are taken from the last intermediate encoder block (layer~23), matching the default setting in \cref{sec:experiments}.
We report mean IoU (mIoU) over randomly sampled episodes.
Because slender structures occupy a tiny fraction of image area, mIoU alone can understate partial recovery of thin branches; we therefore complement \cref{tab:slender_ics_comparison} with layer/resolution sweeps (\cref{tab:slender_layer_fundus,tab:slender_layer_deepglobe}) and layer-wise feature visualizations.

\subsection{Quantitative Comparison on Slender Structures}
\label{sec:slender_quant}

\Cref{tab:slender_ics_comparison} summarizes one-shot and five-shot mIoU on Fundus and DeepGlobe-18.
Three observations stand out.

\paragraph{All methods remain far below compact-object performance.}
Even the best entry in \cref{tab:slender_ics_comparison}---\ours{} at 21.5\% mIoU on Fundus (5-shot)---is an order of magnitude lower than scores on benchmarks such as COCO or Chest X-ray in \cref{tab:main_comparison}.
This confirms that current VFM representations, together with prototype-based ICS pipelines, are insufficient for reliable slender-structure recovery despite strong performance on semantically compact targets.

\paragraph{Relative ranking is preserved, but absolute gains are modest.}
As shown in \cref{tab:slender_ics_comparison}, \ours{} consistently outperforms INSID3 and GF-SAM on both datasets (+7.4/+9.8 pts.\ on Fundus and +6.1/+6.2 pts.\ on DeepGlobe-18 under 1-shot).
Progressive refinement in \ours{} mitigates background-dominated matching to some extent, yet the low performance ceiling suggests that post-hoc score refinement cannot fully compensate for weak patch-level vessel semantics.

\paragraph{Additional references provide limited benefit.}
\cref{tab:slender_ics_comparison} further shows that increasing shots from 1 to 5 improves Fundus mIoU by only 1.8 pts.\ for \ours{} and can even \emph{reduce} INSID3 performance (12.3 $\rightarrow$ 11.7).
On DeepGlobe-18, all methods gain less than 1.5 pts.\ with five references.
This supports the dispersed-feature limitation above: extra prototypes are insufficient to cover the full appearance and topology of elongated structures.

\begin{table}[t]
\centering
\caption{Comparison of training-free ICS methods on slender-structure datasets (mIoU, \%). All methods use frozen DINOv3-L at $1024{\times}1024$ with default last-layer features.}
\label{tab:slender_ics_comparison}
\vspace{-2mm}
\setlength{\tabcolsep}{5pt}
\renewcommand{\arraystretch}{1.15}
\small
\begin{tabular}{l|cc|cc}
\toprule
\multirow{2}{*}{\textbf{Method}}
& \multicolumn{2}{c|}{\textbf{Fundus}}
& \multicolumn{2}{c}{\textbf{DeepGlobe-18}} \\
\cmidrule(lr){2-3}\cmidrule(lr){4-5}
& \textbf{1-shot} & \textbf{5-shot}
& \textbf{1-shot} & \textbf{5-shot} \\
\midrule
GF-SAM~\cite{Zhang:2024:GF-SAM}
& 8.6 & 8.4 & 8.5 & 9.3 \\
INSID3~\cite{cuttano2026insid3}
& 12.3 & 11.7 & 7.2 & 7.3 \\
\textbf{\ours{}}
& \textbf{19.7} & \textbf{21.5} & \textbf{13.3} & \textbf{13.5} \\
\bottomrule
\end{tabular}
\vspace{-2mm}
\end{table}

\subsection{Layer-wise and Resolution Sensitivity}
\label{sec:slender_layer}

\begin{figure}[htbp]
    \centering
    \includegraphics[width=\linewidth]{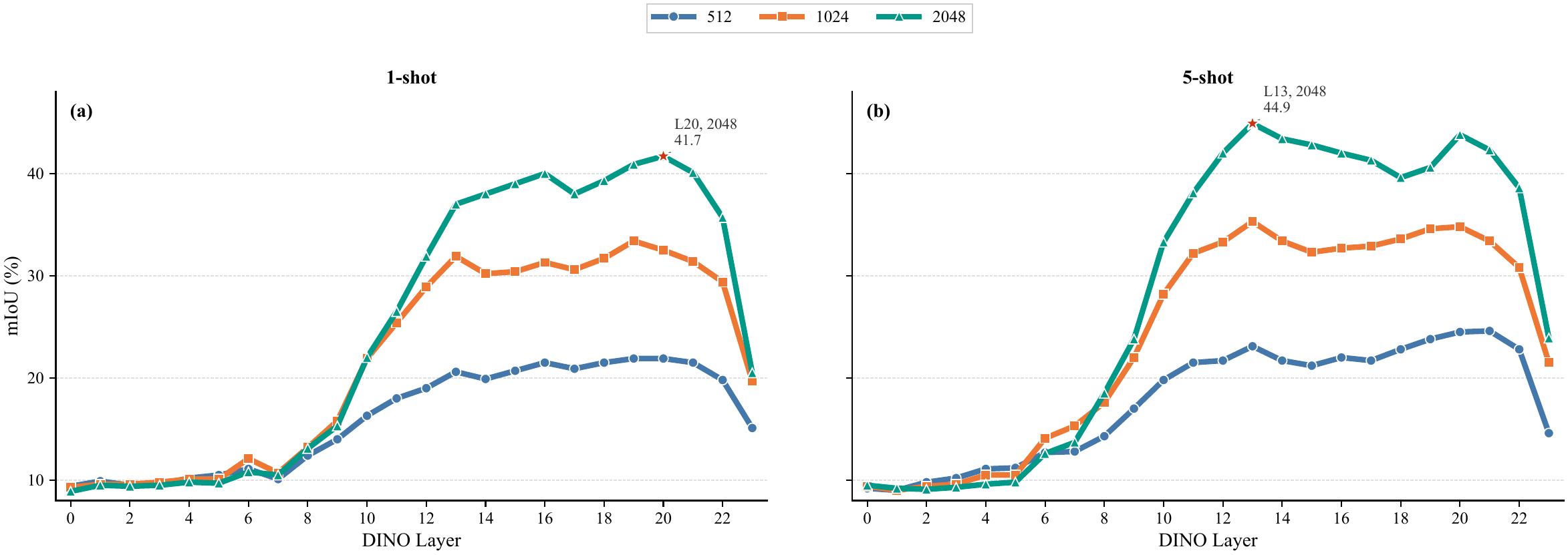}
    \vspace{-1.75em}
    \caption{\textbf{Layer-wise mIoU on Fundus.}
    Curves summarize \cref{tab:slender_layer_fundus}.
    Mid-to-deep layers (8--21) benefit strongly from higher input resolution, whereas the last layer (23) degrades sharply, confirming that the default feature depth is misaligned with slender-structure matching.}
    \label{fig:fundus_layer_iou}
    \vspace{-0.6em}
\end{figure}

\begin{figure}[htbp]
    \centering
    \includegraphics[width=\linewidth]{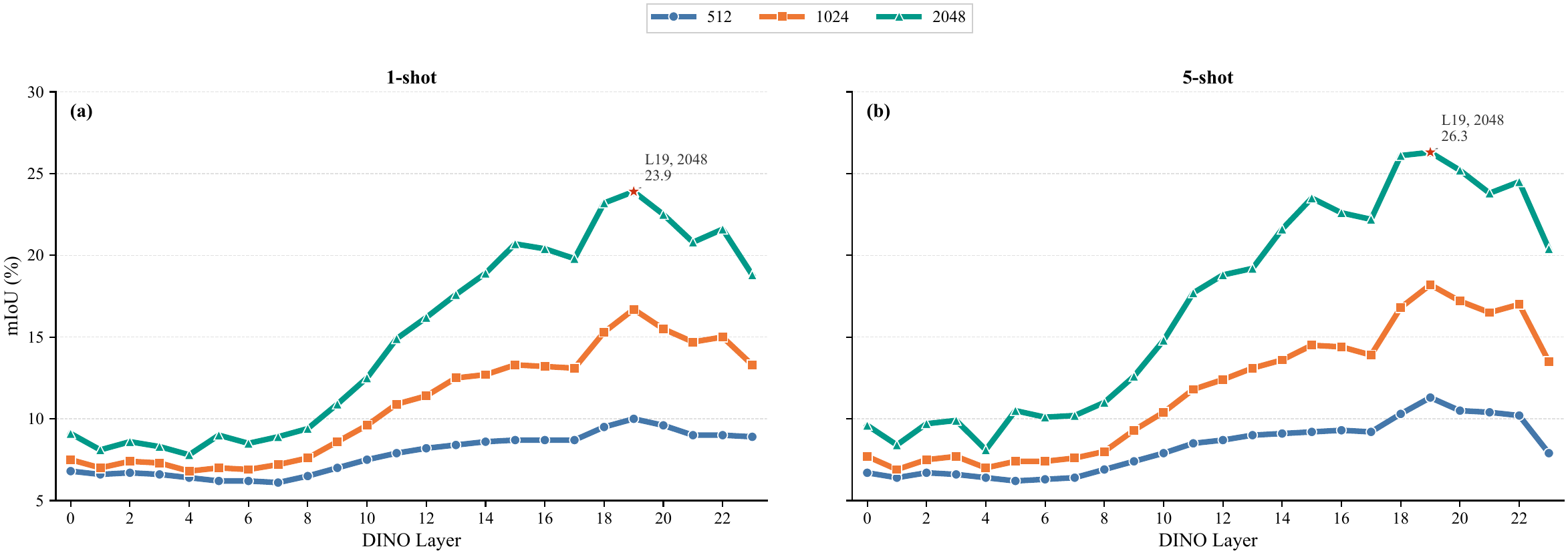}
    \vspace{-1.75em}
    \caption{\textbf{Layer-wise mIoU on DeepGlobe-18.}
    Curves summarize \cref{tab:slender_layer_deepglobe}.
    The pattern mirrors Fundus: mid-to-deep layers gain from higher resolution, while the last layer (23) underperforms, indicating that default last-layer features are poorly suited to elongated targets.}
    \label{fig:deepglobe_layer_iou}
    \vspace{-0.6em}
\end{figure}

Slender structures are highly sensitive to both \emph{which} encoder layer is used and \emph{how} densely patches are sampled.
\Cref{tab:slender_layer_fundus,tab:slender_layer_deepglobe} and \cref{fig:fundus_layer_iou,fig:deepglobe_layer_iou} report layer-wise sweeps on Fundus and DeepGlobe-18 under 1-shot and 5-shot settings with input resolutions of 512, 1024, and 2048.
The same trends appear on both datasets.

\paragraph{Early layers carry little segmentation signal.}
In \cref{tab:slender_layer_fundus,tab:slender_layer_deepglobe}, layers 0--7 yield mIoU below 13\% on Fundus and below 10\% on DeepGlobe-18 across all resolutions and shot counts.
Low-level tokens therefore encode texture and local appearance rather than vessel- or road-level semantics usable for cross-image matching.

\paragraph{Mid-to-deep layers encode partial target semantics.}
Performance rises sharply between layers 8 and 16 in both tables.
On Fundus (\cref{tab:slender_layer_fundus}), 1-shot mIoU reaches 28.9--31.9\% at layer~12 with 2048 inputs; on DeepGlobe-18 (\cref{tab:slender_layer_deepglobe}), the corresponding peak at layer~19 reaches 23.9\%.
Intermediate representations thus begin to capture structure-relevant cues, but only after sufficient depth and spatial resolution.

\paragraph{The default last layer is suboptimal.}
Layer~23---used by default in our main pipeline and in \cref{tab:slender_ics_comparison}---achieves only 19.7\% mIoU on Fundus and 13.3\% on DeepGlobe-18 at 1024 (1-shot), as reported in \cref{tab:slender_layer_fundus,tab:slender_layer_deepglobe}.
In contrast, the best layer at the same resolution (layer~20 on Fundus, layer~19 on DeepGlobe-18) and the best overall settings (layers~20--21 at 2048 on Fundus, up to 41.7\%; layer~19 at 2048 on DeepGlobe-18, up to 23.9\%) are substantially higher.
The final block appears over-specialized for global semantic aggregation and sacrifices the fine spatial detail needed for thin structures, consistent with the structural-preservation limitation above.

\paragraph{Resolution is critical for thin structures.}
At the best-performing mid-to-deep layers in \cref{tab:slender_layer_fundus,tab:slender_layer_deepglobe}, increasing resolution from 512 to 2048 nearly doubles 1-shot mIoU on Fundus (21.9\% $\rightarrow$ 41.7\% at layer~20) and more than doubles it on DeepGlobe-18 (10.0\% $\rightarrow$ 23.9\% at layer~19).
Higher resolution reduces foreground--background mixing within each patch token and lets cross-image matching operate on a finer spatial grid, directly mitigating the patch-mixing issue described above.
Even so, the best layer--resolution combinations in \cref{tab:slender_layer_fundus,tab:slender_layer_deepglobe} remain below 45\% mIoU on Fundus and 27\% on DeepGlobe-18, indicating that representation and matching improvements alone are insufficient without explicit structural priors.

\begin{table}[htbp]
\centering
\caption{Layer-wise mIoU (\%) on Fundus with \ours{} under 1-shot and 5-shot settings and input resolutions of 512, 1024, and 2048. Bold indicates the best result per column.}
\label{tab:slender_layer_fundus}
\vspace{-2mm}
\setlength{\tabcolsep}{3.5pt}
\renewcommand{\arraystretch}{1.15}
\small
\begin{tabular}{c|ccc|ccc}
\toprule
\multirow{2}{*}{\textbf{Layer}}
& \multicolumn{3}{c|}{\textbf{1-shot}}
& \multicolumn{3}{c}{\textbf{5-shot}} \\
\cmidrule(lr){2-4}\cmidrule(l){5-7}
& \textbf{512} & \textbf{1024} & \textbf{2048}
& \textbf{512} & \textbf{1024} & \textbf{2048} \\
\midrule
0  & 9.4  & 9.3  & 8.9  & 9.2  & 9.4  & 9.5  \\
1  & 9.9  & 9.6  & 9.5  & 9.0  & 9.0  & 9.2  \\
2  & 9.5  & 9.6  & 9.4  & 9.8  & 9.4  & 9.1  \\
3  & 9.6  & 9.8  & 9.5  & 10.2 & 9.6  & 9.3  \\
4  & 10.2 & 10.1 & 9.8  & 11.1 & 10.5 & 9.6  \\
5  & 10.5 & 10.1 & 9.7  & 11.2 & 10.5 & 9.8  \\
6  & 11.1 & 12.1 & 10.8 & 12.7 & 14.1 & 12.6 \\
7  & 10.1 & 10.7 & 10.5 & 12.8 & 15.3 & 13.7 \\
8  & 12.4 & 13.2 & 13.1 & 14.3 & 17.6 & 18.5 \\
9  & 14.0 & 15.8 & 15.3 & 17.0 & 22.0 & 23.8 \\
10 & 16.3 & 21.9 & 22.0 & 19.8 & 28.2 & 33.3 \\
11 & 18.0 & 25.4 & 26.5 & 21.5 & 32.2 & 38.1 \\
12 & 19.0 & 28.9 & 31.9 & 21.7 & 33.3 & 42.0 \\
13 & 20.6 & 31.9 & 37.0 & 23.1 & \textbf{35.3} & \textbf{44.9} \\
14 & 19.9 & 30.2 & 38.0 & 21.7 & 33.4 & 43.4 \\
15 & 20.7 & 30.4 & 39.0 & 21.2 & 32.3 & 42.8 \\
16 & 21.5 & 31.3 & 40.0 & 22.0 & 32.7 & 42.0 \\
17 & 20.9 & 30.6 & 38.0 & 21.7 & 32.9 & 41.3 \\
18 & 21.5 & 31.7 & 39.3 & 22.8 & 33.6 & 39.6 \\
19 & \textbf{21.9} & \textbf{33.4} & 40.9 & 23.8 & 34.6 & 40.6 \\
20 & \textbf{21.9} & 32.5 & \textbf{41.7} & 24.5 & 34.8 & 43.8 \\
21 & \textbf{21.9} & 31.4 & 40.1 & \textbf{24.6} & 33.4 & 42.3 \\
22 & 19.8 & 29.4 & 35.7 & 22.8 & 30.8 & 38.6 \\
23 & 15.1 & 19.7 & 20.5 & 14.6 & 21.5 & 23.9 \\
\bottomrule
\end{tabular}
\vspace{-2mm}
\end{table}

\begin{table}[htbp]
\centering
\caption{Layer-wise mIoU (\%) on DeepGlobe-18 with \ours{} under 1-shot and 5-shot settings and input resolutions of 512, 1024, and 2048. Bold indicates the best result per column.}
\label{tab:slender_layer_deepglobe}
\vspace{-2mm}
\setlength{\tabcolsep}{3.5pt}
\renewcommand{\arraystretch}{1.15}
\small
\begin{tabular}{c|ccc|ccc}
\toprule
\multirow{2}{*}{\textbf{Layer}}
& \multicolumn{3}{c|}{\textbf{1-shot}}
& \multicolumn{3}{c}{\textbf{5-shot}} \\
\cmidrule(lr){2-4}\cmidrule(l){5-7}
& \textbf{512} & \textbf{1024} & \textbf{2048}
& \textbf{512} & \textbf{1024} & \textbf{2048} \\
\midrule
0  &  6.8 &  7.5 &  9.1 &  6.7 &  7.7 &  9.6 \\
1  &  6.6 &  7.0 &  8.1 &  6.4 &  6.9 &  8.4 \\
2  &  6.7 &  7.4 &  8.6 &  6.7 &  7.5 &  9.7 \\
3  &  6.6 &  7.3 &  8.3 &  6.6 &  7.7 &  9.9 \\
4  &  6.4 &  6.8 &  7.8 &  6.4 &  7.0 &  8.1 \\
5  &  6.2 &  7.0 &  9.0 &  6.2 &  7.4 & 10.5 \\
6  &  6.2 &  6.9 &  8.5 &  6.3 &  7.4 & 10.1 \\
7  &  6.1 &  7.2 &  8.9 &  6.4 &  7.6 & 10.2 \\
8  &  6.5 &  7.6 &  9.4 &  6.9 &  8.0 & 11.0 \\
9  &  7.0 &  8.6 & 10.9 &  7.4 &  9.3 & 12.6 \\
10 &  7.5 &  9.6 & 12.5 &  7.9 & 10.4 & 14.8 \\
11 &  7.9 & 10.9 & 14.9 &  8.5 & 11.8 & 17.7 \\
12 &  8.2 & 11.4 & 16.2 &  8.7 & 12.4 & 18.8 \\
13 &  8.4 & 12.5 & 17.6 &  9.0 & 13.1 & 19.2 \\
14 &  8.6 & 12.7 & 18.9 &  9.1 & 13.6 & 21.6 \\
15 &  8.7 & 13.3 & 20.7 &  9.2 & 14.5 & 23.5 \\
16 &  8.7 & 13.2 & 20.4 &  9.3 & 14.4 & 22.6 \\
17 &  8.7 & 13.1 & 19.8 &  9.2 & 13.9 & 22.2 \\
18 &  9.5 & 15.3 & 23.2 & 10.3 & 16.8 & 26.1 \\
19 & \textbf{10.0} & \textbf{16.7} & \textbf{23.9} & \textbf{11.3} & \textbf{18.2} & \textbf{26.3} \\
20 &  9.6 & 15.5 & 22.5 & 10.5 & 17.2 & 25.2 \\
21 &  9.0 & 14.7 & 20.8 & 10.4 & 16.5 & 23.8 \\
22 &  9.0 & 15.0 & 21.6 & 10.2 & 17.0 & 24.5 \\
23 &  8.9 & 13.3 & 18.8 &  7.9 & 13.5 & 20.4 \\
\bottomrule
\end{tabular}
\vspace{-2mm}
\end{table}

\subsection{Feature Representation Analysis}
\label{sec:slender_feature}

\begin{figure}[t]
    \centering
    \includegraphics[width=\linewidth]{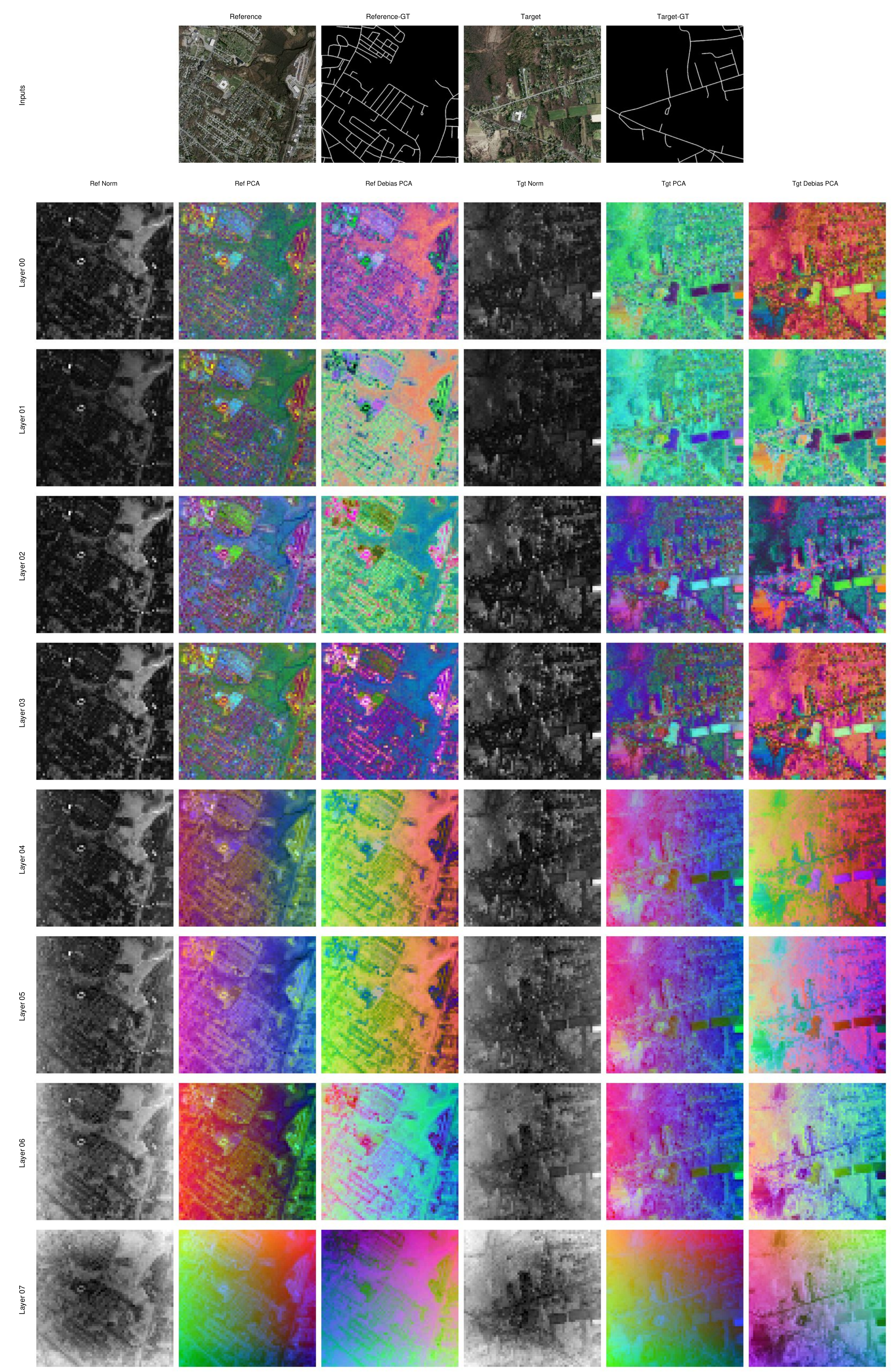}
    \vspace{-1.75em}
    \caption{\textbf{Layer-wise DINOv3 features on DeepGlobe-18 (layers 0--7).}
    Top row: reference image, reference mask, target image, and ground truth.
    Each subsequent row shows, for reference \emph{(left)} and target \emph{(right)}, the feature norm (\textbf{Norm}), PCA colorization (\textbf{PCA}), and debiased PCA (\textbf{Debias PCA}).}
    \label{fig:deepglobe_feat_0_7}
\end{figure}

\begin{figure}[t]
    \centering
    \includegraphics[width=\linewidth]{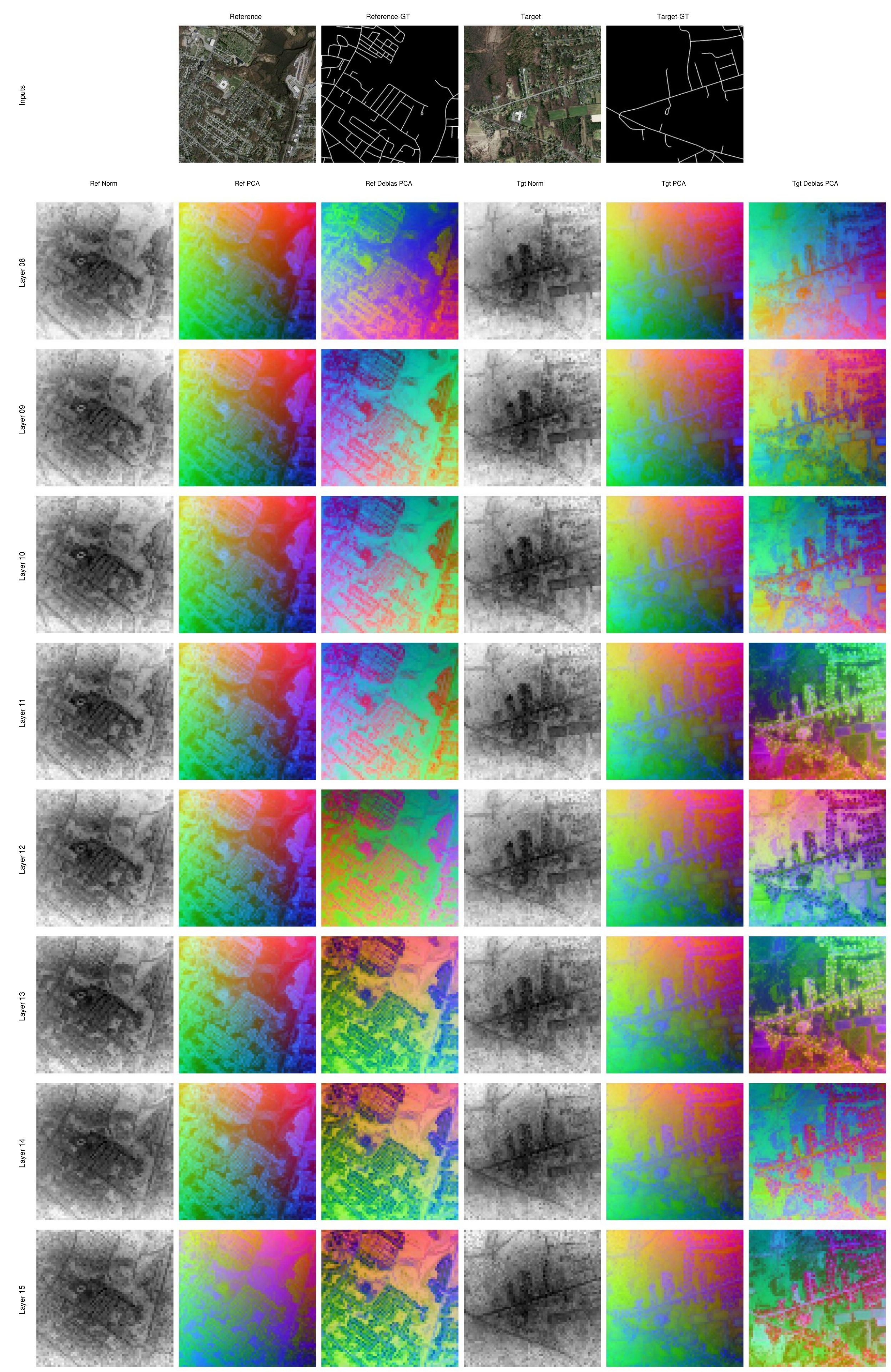}
    \vspace{-1.75em}
    \caption{\textbf{Layer-wise DINOv3 features on DeepGlobe-18 (layers 8--15).}
    Same layout as \cref{fig:deepglobe_feat_0_7}.}
    \label{fig:deepglobe_feat_8_15}
\end{figure}

\begin{figure}[t]
    \centering
    \includegraphics[width=\linewidth]{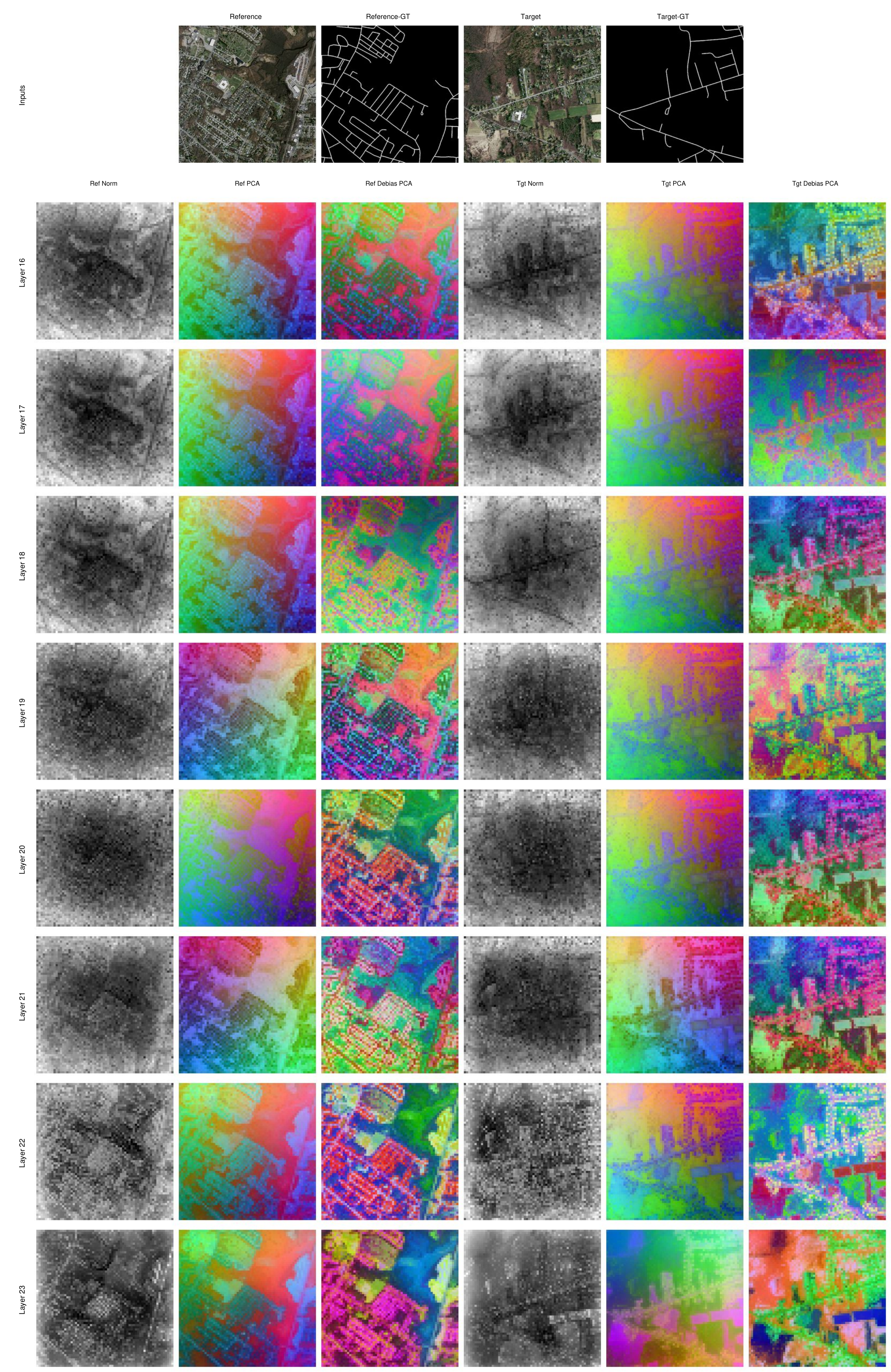}
    \vspace{-1.75em}
    \caption{\textbf{Layer-wise DINOv3 features on DeepGlobe-18 (layers 16--23).}
    Same layout as \cref{fig:deepglobe_feat_0_7}.}
    \label{fig:deepglobe_feat_16_23}
\end{figure}

\begin{figure}[t]
    \centering
    \includegraphics[width=\linewidth]{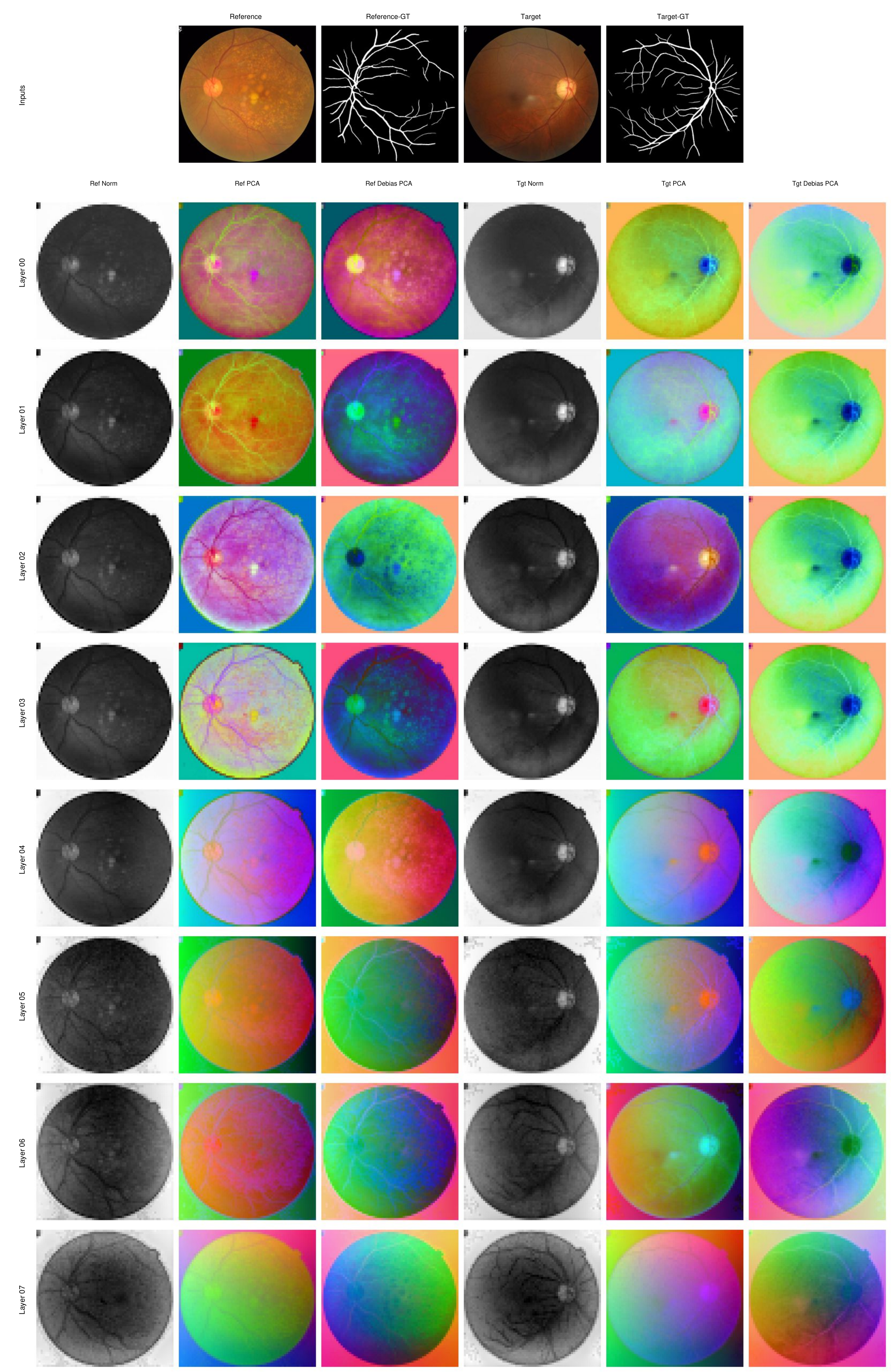}
    \vspace{-1.75em}
    \caption{\textbf{Layer-wise DINOv3 features on Fundus (layers 0--7).}
    Same layout as \cref{fig:deepglobe_feat_0_7}.}
    \label{fig:fundus_feat_0_7}
\end{figure}

\begin{figure}[t]
    \centering
    \includegraphics[width=\linewidth]{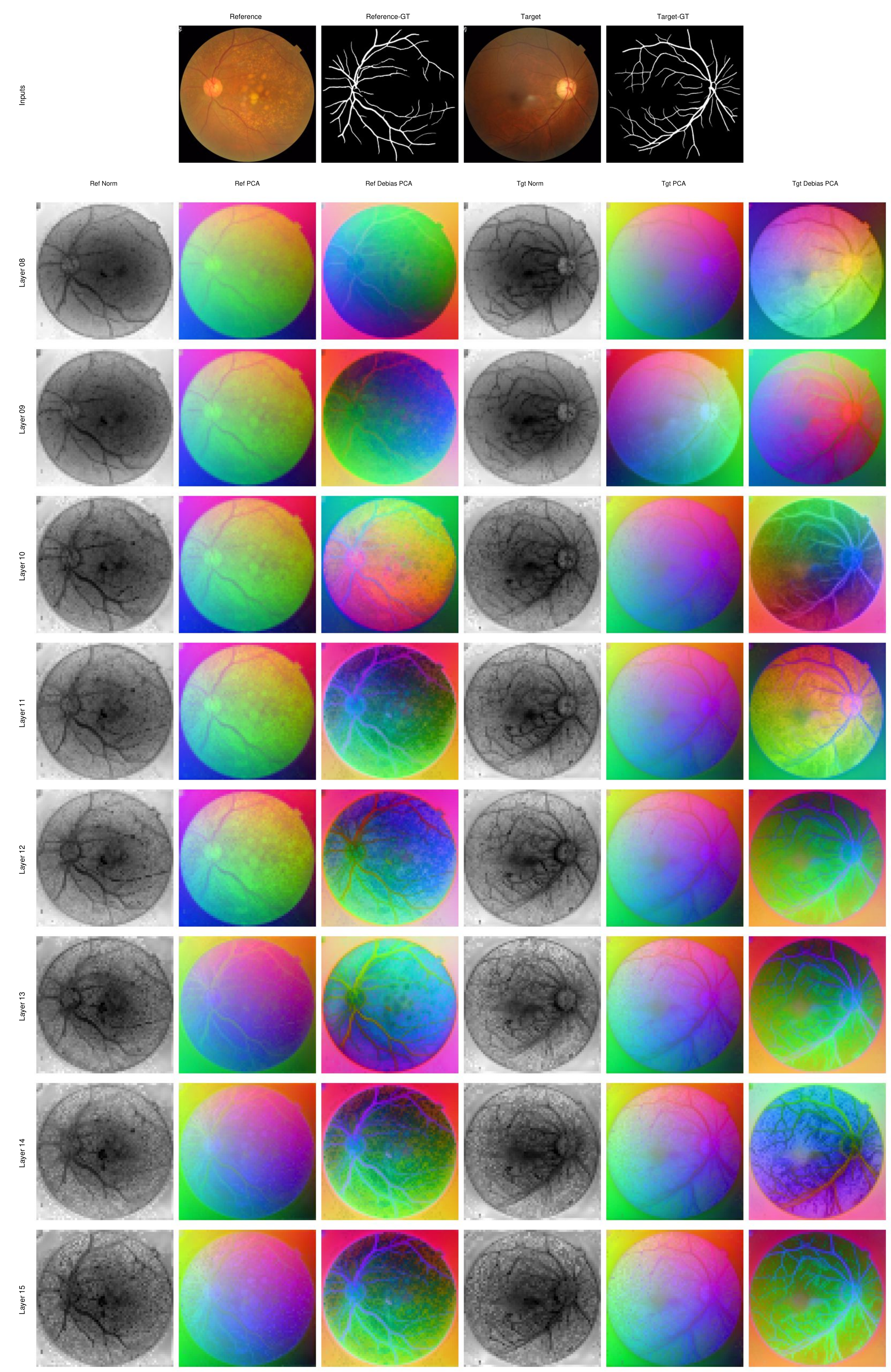}
    \vspace{-1.75em}
    \caption{\textbf{Layer-wise DINOv3 features on Fundus (layers 8--15).}
    Same layout as \cref{fig:deepglobe_feat_0_7}.}
    \label{fig:fundus_feat_8_15}
\end{figure}

\begin{figure}[t]
    \centering
    \includegraphics[width=\linewidth]{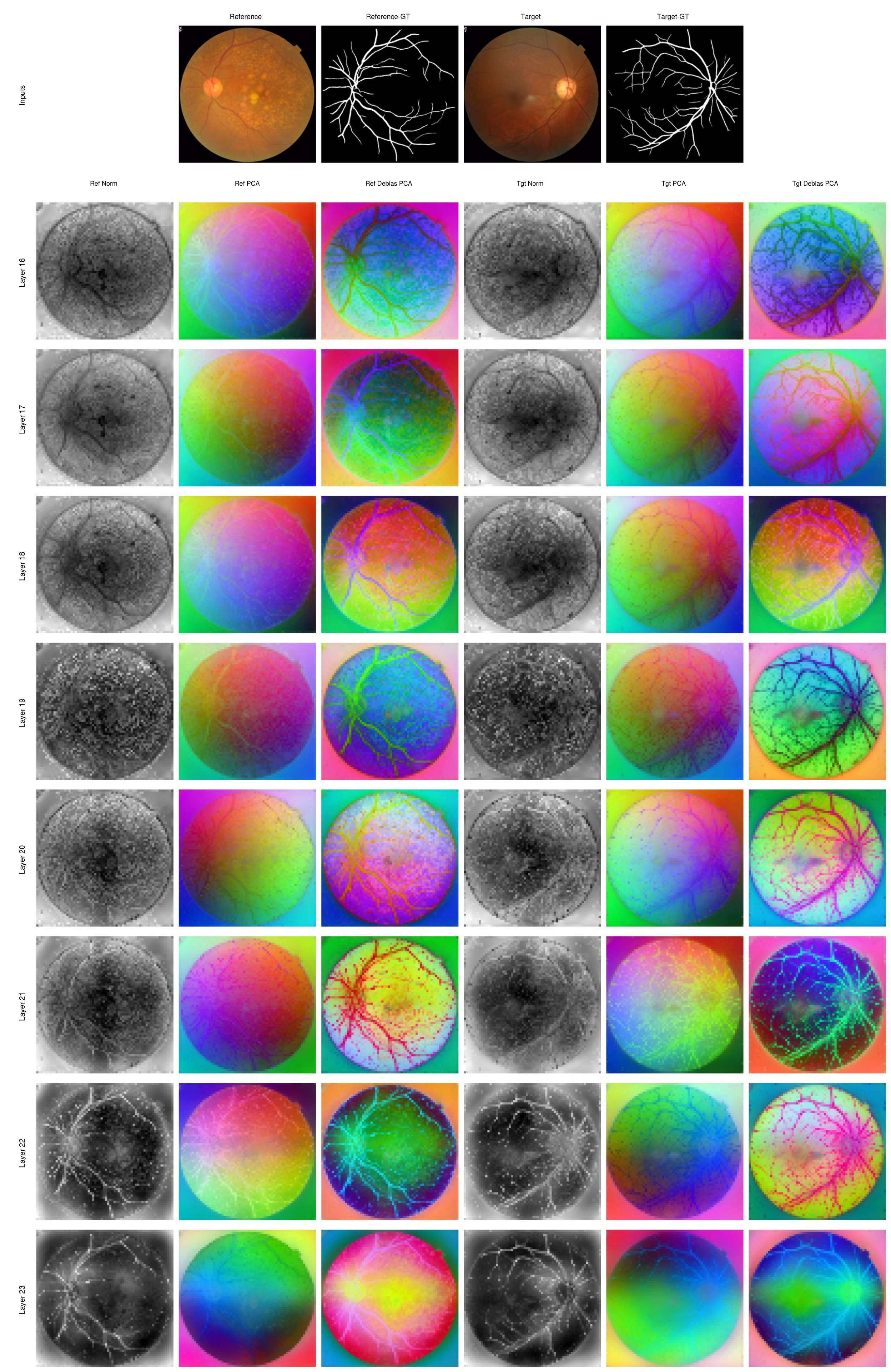}
    \vspace{-1.75em}
    \caption{\textbf{Layer-wise DINOv3 features on Fundus (layers 16--23).}
    Same layout as \cref{fig:deepglobe_feat_0_7}.}
    \label{fig:fundus_feat_16_23}
\end{figure}

To connect the quantitative gap in \cref{tab:slender_ics_comparison,tab:slender_layer_fundus,tab:slender_layer_deepglobe} to representation behavior, we visualize DINOv3 patch features layer by layer on representative Fundus and DeepGlobe-18 episodes (\cref{fig:deepglobe_feat_0_7,fig:deepglobe_feat_8_15,fig:deepglobe_feat_16_23,fig:fundus_feat_0_7,fig:fundus_feat_8_15,fig:fundus_feat_16_23}).
For each encoder block we show the feature norm, a PCA colorization of raw tokens, and the same PCA view after positional debiasing.

\paragraph{Foreground--background mixing in patch tokens.}
In early and mid layers, PCA maps on vessel or road pixels closely resemble surrounding background: token colors vary smoothly across tissue or terrain boundaries instead of forming a coherent foreground cluster.
Norm maps further show that high-magnitude responses often align with global illumination or texture rather than with thin foreground branches.
A single token therefore rarely represents ``vessel'' or ``road'' in isolation, directly illustrating the patch-mixing limitation.

\paragraph{Dispersed foreground semantics across branches.}
Even in deeper layers where foreground regions become slightly more distinguishable, different branches, crossings, and terminators appear as separate color modes in PCA space rather than as one compact cluster.
This dispersion explains why prototype-based ICS---which compresses the reference mask into a small set of cluster centers---systematically misses thin side branches and peripheral capillaries.

\paragraph{Positional debiasing improves alignment but not topology.}
Positional debiasing (Debias PCA) reduces layout-driven color gradients and sharpens cross-image correspondence, especially when reference and target objects occupy different locations.
It does not, however, reconnect fragmented segments in similarity maps: debiasing improves \emph{where} to match but not \emph{whether} matched regions form a connected structure.
This aligns with the modest quantitative gains of \ours{} over INSID3---purification improves semantic alignment, yet structural continuity remains largely unaddressed.

\end{document}